%% file: main.tex
\documentclass[letterpaper]{article} 
\usepackage[preprint]{aaai2027}
\usepackage[hyphens]{url}  
\usepackage{graphicx} 
\usepackage{natbib}  
\usepackage{caption} 
\usepackage{amsmath}
\usepackage{amssymb}
\usepackage{amsthm}
\usepackage{algorithm}
\usepackage{algorithmic}
\usepackage{booktabs}
\usepackage{multirow}

\title{MedAD-R1: Consistency-Reinforced Policy Optimization for Interpretable Medical Anomaly Detection}
\author{
Haitao Zhang$^{1}$\thanks{Equal contribution.},
Yingying Wang$^{1,2,\ast}$,
Jiaxiang Wang$^{1,2}$,
Haote Xu$^{3,4}$\\
Hongyang Zhang$^{5}$,
Yirong Chen$^{6}$,
Yue Huang$^{1,2}$,
Xinghao Ding$^{1,2}$\thanks{Corresponding author.}
}
\affiliations{
$^{1}$School of Informatics, Xiamen University \quad $^{2}$Institute of Artificial Intelligence, Xiamen University\\
$^{3}$Zhejiang Expressway Co., Ltd. \quad $^{4}$School of Transportation Science and Engineering, Beihang University\\
$^{5}$School of Science and Engineering, Chinese University of Hong Kong \quad $^{6}$Shanghai Artificial Intelligence Laboratory\\
haitaozhang0829@gmail.com, dxh@xmu.edu.cn
}

\begin{document}

\maketitle

\begin{abstract}
Medical Anomaly Detection (MedAD) offers a promising direction for medical image analysis with Large Multimodal Models (LMMs). However, progress is limited by fragmented datasets and the tendency of Supervised Fine-Tuning (SFT) to learn superficial image-text correlations rather than verifiable diagnostic reasoning. Consequently, current models often generate fluent explanations that are either insufficiently grounded in the image or inconsistent with their final answers, limiting their reliability in high-stakes medical applications. To address these issues, we introduce MedAD-38K, a large-scale, multimodal, and multicenter benchmark containing structured Visual Question Answering pairs and quality-controlled diagnostic Chain-of-Thought annotations across five core MedAD tasks. Based on this benchmark, we propose a two-stage framework. Cognitive Injection first uses SFT to inject domain-specific medical knowledge and establish a structured think-then-answer format. Consistency Group Relative Policy Optimization (Con-GRPO) then employs an Evidence-Aware Consistency Reward to reinforce reasoning that remains grounded in the image and logically supports the final answer. The resulting MedAD-R1 achieves state-of-the-art performance on MedAD-38K and consistently outperforms all evaluated baselines across five source-disjoint external datasets spanning diverse imaging modalities. Beyond accuracy, it achieves higher reasoning-answer consistency and visual-grounding scores across multiple backbones. With only 0.8B parameters, MedAD-R1 offers practical potential for resource-constrained deployment. These results demonstrate the generality of evidence-aware consistency optimization for interpretable MedAD. Project resources are available at \pdfstartlink attr {/Border [0 0 0]} user {/Subtype /Link /A << /S /URI /URI (https://github.com/zhtstar/MedAD-R1) >>}\url{https://github.com/zhtstar/MedAD-R1}\pdfendlink.
\end{abstract}

\section{Introduction}

\begin{figure}[t]
    \centering
    \includegraphics[width=\columnwidth]{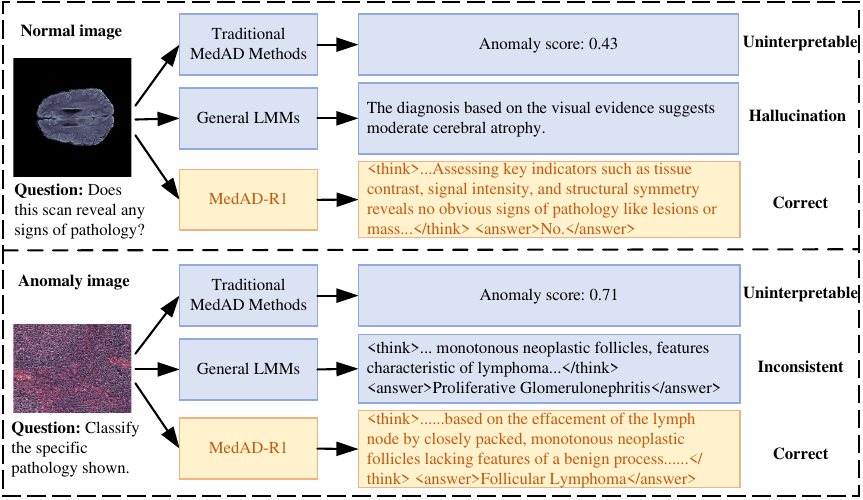}
    \caption{Comparison of traditional MedAD methods, general-purpose LMMs, and MedAD-R1. General LMMs may hallucinate findings (top) or produce reasoning inconsistent with the final answer (bottom), whereas MedAD-R1 remains visually grounded and answer-consistent.}
    \label{fig:intro_comparison}
\end{figure}

Medical Anomaly Detection (MedAD) plays an important role in identifying abnormal patterns that may indicate disease or pathological change in medical images. Traditional MedAD methods usually produce anomaly scores or localization maps, but they offer limited explanatory context~\cite{deng2024simclip}. Large Multimodal Models (LMMs) provide a promising alternative by combining visual understanding with natural-language reasoning. Their explanations, however, are not always reliable. As illustrated in Figure~\ref{fig:intro_comparison}, two failure modes are particularly important. First, a model may describe pathological findings that are not supported by the image. Second, its reasoning may not logically support its final answer. Both failures break the image-reasoning-answer chain and limit the reliability of LMMs in high-stakes medical applications.

Two factors underlie these failures. The first factor concerns the available data resources. Existing MedAD datasets are fragmented across imaging modalities, anatomical regions, and diagnostic tasks, which makes it difficult to train and fairly evaluate models for robust generalization~\cite{prevedello2019challenges}. Most datasets provide only image-level labels, which cannot supervise or assess structured diagnostic reasoning. The second factor concerns the training objective used to adapt medical LMMs. Many medical LMMs, including LLaVA-Med~\cite{llava-med} and HuatuoGPT-Vision~\cite{huatuogpt}, are trained primarily through Supervised Fine-Tuning (SFT), which optimizes the likelihood of reference responses. However, standard SFT does not explicitly evaluate whether the generated reasoning is grounded in the image or logically supports the stated answer~\cite{gudibande2023false}. Consequently, neither failure mode is directly penalized during training. A more suitable objective should therefore optimize not only answer accuracy, but also the connection between visual evidence, reasoning, and the final decision.

To address these challenges, we first construct \textbf{MedAD-38K}, a large-scale, multimodal, and multicenter benchmark for MedAD. It contains structured Visual Question Answering pairs and quality-controlled Chain-of-Thought annotations across five core MedAD tasks: anatomy identification, modality classification, anomaly detection, pathology characterization, and lesion localization. To measure generalization beyond the benchmark itself, we further establish a unified zero-shot MedAD evaluation protocol using five source-disjoint external datasets spanning diverse imaging modalities. Building on MedAD-38K, we propose a two-stage training framework that yields our model, \textbf{MedAD-R1}. The first stage, Cognitive Injection, uses SFT to inject domain-specific medical knowledge and to establish a structured think-then-answer format. The second stage introduces \textbf{Consistency Group Relative Policy Optimization (Con-GRPO)}, whose Evidence-Aware Consistency Reward scores the generated reasoning along five dimensions: visual grounding, task relevance, answer support, no-hallucination, and no-contradiction. The reward thus spans the image-to-reasoning link, the reasoning-to-answer link, and the relevance of the reasoning to the diagnostic question, encouraging reasoning that is grounded in the image and aligned with the final answer. We further apply the same framework to multiple Qwen and InternVL backbones to test whether the proposed optimization depends on a particular architecture.

Extensive experiments show that MedAD-R1 achieves SOTA performance on MedAD-38K and consistently outperforms all evaluated baselines across the five source-disjoint external datasets. Beyond answer accuracy, it attains higher visual grounding and reasoning-answer consistency scores, and these gains hold across different backbones. Despite having only 0.8B parameters, MedAD-R1 surpasses substantially larger general-purpose and domain-specific LMMs.

Our contributions are summarized as follows:
\begin{itemize}
\item We construct MedAD-38K, a large-scale, multimodal, and multicenter benchmark with structured VQA pairs and quality-controlled CoT annotations for robust training and systematic evaluation of reasoning in MedAD. We also establish a unified zero-shot evaluation protocol using five source-disjoint external datasets to assess cross-source generalization.

\item We propose a two-stage training framework centered on Con-GRPO. Its Evidence-Aware Consistency Reward extends beyond reasoning-answer agreement by explicitly encouraging the reasoning to be grounded in visual evidence, thereby reducing self-consistent but unsupported explanations. The framework yields consistent improvements across multiple model backbones.

\item We develop MedAD-R1, a lightweight 0.8B model that achieves strong answer accuracy, visual grounding, reasoning-answer consistency, and external generalization. Its compact scale and favorable inference efficiency further support its potential for resource-constrained MedAD applications.

\end{itemize}

\section{Related Work}

\subsection{Medical Anomaly Detection}
Early approaches to Medical Anomaly Detection (MedAD) primarily follow reconstruction-based~\cite{cai2024rethinking,iqbal2023unsupervised} or feature-matching paradigms~\cite{zhang2024mediclip}. These methods typically learn the distribution of normal tissue and identify anomalies as deviations from this distribution. Although effective in many settings, their performance may degrade under changes in imaging protocols, acquisition devices, or data distributions. Their outputs are also usually limited to anomaly scores or localization maps, providing little natural-language explanation for why a region is considered abnormal. Our work complements them by making the explanation itself an object of training and evaluation, requiring it to be grounded in the image and consistent with the final answer.

\subsection{Large Multimodal Models in Medicine}
The development of Large Multimodal Models (LMMs) has encouraged specialized medical variants, since general-purpose models such as GPT-4o~\cite{hurst2024gpt} are not specifically optimized for medical image understanding. Following a common architecture consisting of a vision encoder, a projection module, and an LLM decoder, LLaVA-Med~\cite{llava-med} continues pretraining on biomedical figure-caption pairs, while Med-Flamingo~\cite{moor2023med} introduces cross-attention layers into the Flamingo architecture for medical adaptation. These models can generate fluent medical descriptions, but most rely heavily on supervised pretraining and Supervised Fine-Tuning (SFT). Such objectives improve domain adaptation but do not explicitly verify whether the generated reasoning is supported by the image or aligned with the final answer. Consequently, plausible but unsupported explanations may remain unpenalized. In contrast, our framework introduces an explicit training signal for the reasoning process itself, rewarding rationales that remain grounded in the image and consistent with the final answer.

\subsection{Reasoning via CoT and RL}
Chain-of-Thought (CoT) prompting makes intermediate reasoning steps observable, and recent studies extend this paradigm to multimodal tasks by decomposing visual-language problems into sequential steps~\cite{zhang2023multimodal}. Although frontier models demonstrate strong CoT-style reasoning on natural images~\cite{gemini}, medical image reasoning often depends on subtle pathological findings and anatomical knowledge. The generated rationale must therefore remain closely connected to the visual evidence. Reinforcement learning has recently been applied to medical multimodal reasoning. For example, MedVLM-R1~\cite{medvlm} primarily rewards the correctness of the final answer. Outcome-based rewards can improve answer accuracy, but they do not directly verify the reasoning used to reach that answer. Our Evidence-Aware Consistency Reward instead evaluates the image-to-reasoning and reasoning-to-answer links by considering visual grounding, task relevance, answer support, no-hallucination, and no-contradiction.

\begin{figure*}[t]
    \centering
    \includegraphics[width=\textwidth]{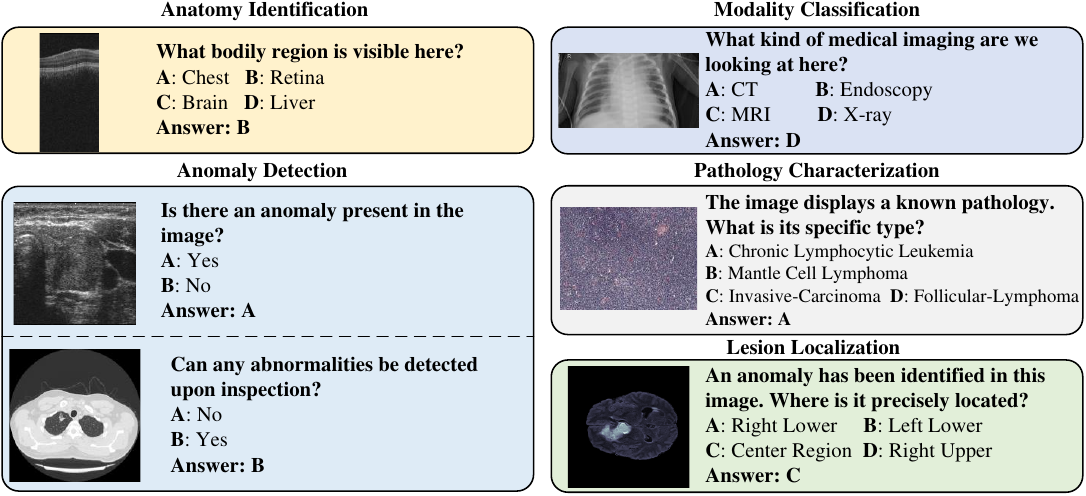}
    \caption{Examples of the five MedAD VQA tasks in MedAD-38K. Questions are formulated as multiple-choice problems to support standardized evaluation.}
    \label{fig:vqa_tasks}
\end{figure*}

\section{Methodology}

We first describe the construction of MedAD-38K, and then present the two-stage framework that turns a general-purpose LMM into MedAD-R1.

\subsection{The MedAD-38K Benchmark}
\label{sec:dataset}

Existing MedAD datasets are often restricted to a single imaging modality, anatomical region, or diagnostic task, and most provide only image-level labels. Such resources are insufficient for training and evaluating structured medical reasoning. We therefore construct \textbf{MedAD-38K} by aggregating and re-annotating 38,422 medical images from 16 public datasets~\cite{braints2021,RESC,al2020dataset,codella2019skin,morozov2020mosmeddata,liew2022large,pedraza2015open,jha2019kvasir,bernal2015wm,mendoncca2013ph,porwal2018indian,popa2025pedidemi,hamada2020br35h,wang2017chestx,orlov2010automatic,aresta2019bach}. The benchmark covers 10 imaging modalities and 10 anatomical regions. Detailed source-, modality-, and task-level statistics are provided in supplementary Section~\ref{app:composition}.

Each image is paired with VQA instances according to the annotations available in its source dataset. The questions cover five MedAD tasks: \textit{Anatomy Identification}, \textit{Modality Classification}, \textit{Anomaly Detection}, \textit{Pathology Characterization}, and \textit{Lesion Localization}, as illustrated in Figure~\ref{fig:vqa_tasks}. For each applicable task, the question is sampled from ten semantically equivalent templates, and answer options are randomly ordered. Anomaly detection uses two options, whereas the other four tasks use four. All correct answers come directly from source labels, masks, or metadata rather than LMMs; full construction details are in supplementary Section~\ref{app:vqa_construction}.

\begin{figure}[t]
    \centering
    \includegraphics[width=0.4\columnwidth]{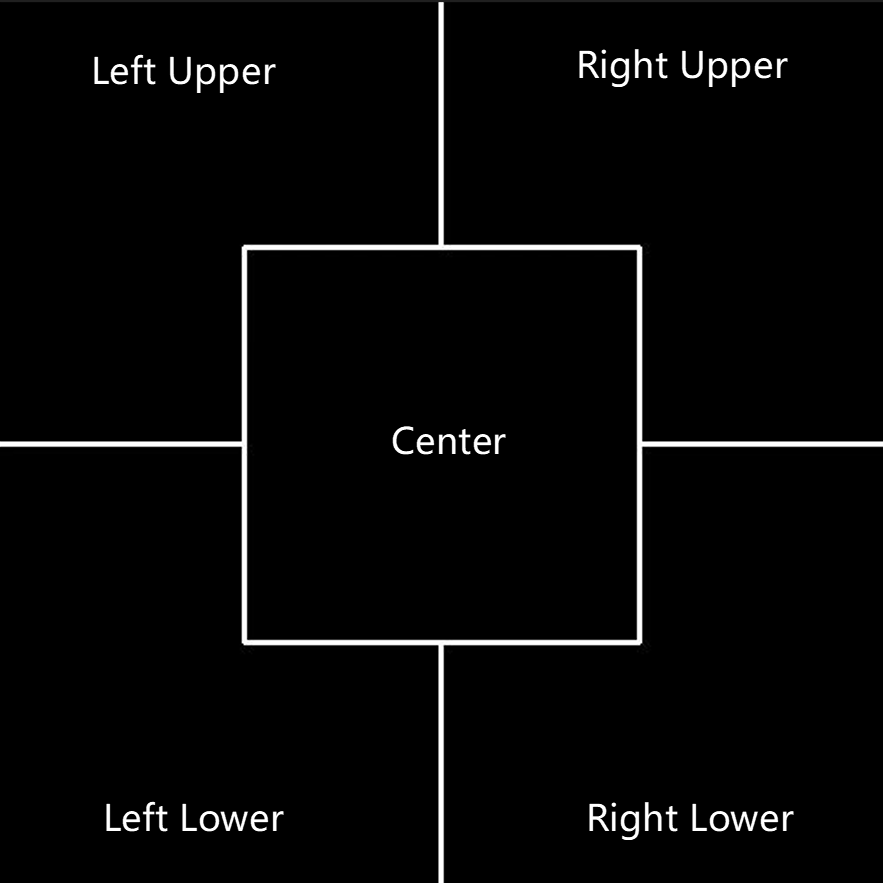}
    \caption{The equal-area five-region spatial grid used to derive lesion-localization labels in MedAD-38K.}
    \label{fig:localization_grid}
\end{figure}

For lesion localization, we map each anomaly mask to one of the five spatial regions illustrated in Figure~\ref{fig:localization_grid}. The label is assigned to the region containing the largest number of anomalous pixels. The motivation, coordinate definitions, and handling of boundary cases are described in supplementary Section~\ref{app:five_region}.

\begin{figure*}[t]
    \centering
    \includegraphics[width=\textwidth]{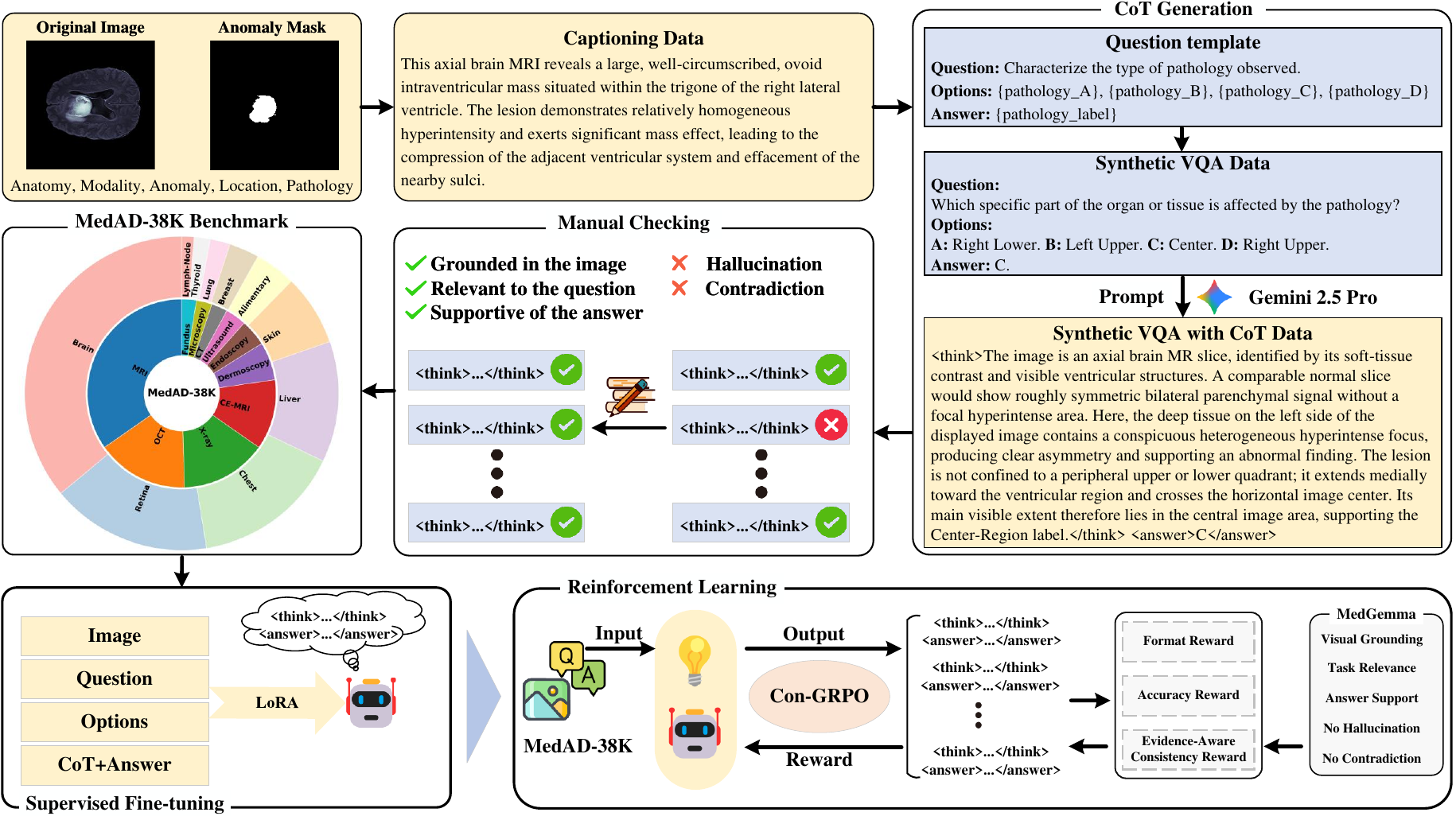}
    \caption{Overview of the MedAD-38K construction and MedAD-R1 training pipeline. Top panel: source labels, masks, and metadata determine the VQA answers. MedGemma-27B generates auxiliary image descriptions, Gemini 2.5 Pro drafts the CoT annotations, and human reviewers verify their quality. Bottom panel: Cognitive Injection first adapts the model through SFT, followed by Con-GRPO with an Evidence-Aware Consistency Reward scored by MedGemma-4B along five dimensions: visual grounding (VG), task relevance (TR), answer support (AS), no-hallucination (NH), and no-contradiction (NC).}
    \label{fig:framework}
\end{figure*}

The diagnostic CoT annotations are generated through the pipeline shown in the top panel of Figure~\ref{fig:framework}. MedGemma~\cite{medgemma} first generates an auxiliary description of the medical image. The image, question, answer options, source-derived answer, and auxiliary description are then provided to Gemini 2.5 Pro~\cite{gemini} to generate an initial reasoning trace. The ground-truth answer remains fixed throughout this process. Each generated trace is subsequently reviewed to determine whether it is grounded in the image, relevant to the question, supportive of the answer, and free from hallucination or contradiction. Invalid traces are revised or rejected and regenerated. This process reduces unsupported and inconsistent reasoning in the supervision data. The full annotation and quality-control procedures are reported in supplementary Sections~\ref{app:cot_generation} and~\ref{app:quality_control}.


\subsection{The MedAD-R1 Framework}
\label{sec:framework}
The architecture of our proposed model, MedAD-R1, is designed to process multimodal inputs and generate a structured, two-part output comprising a diagnostic reasoning trace and a final answer. As illustrated in Figure~\ref{fig:framework}, the model takes a medical image $I$ and a corresponding textual question $Q$ as input. It then generates a structured output $Y$, which consists of a CoT text, denoted as the thought $T$, and a final diagnostic answer $A$.

\subsection{Cognitive Injection via SFT}
\label{sec:sft}

The primary goal of the first stage is twofold: (i) to inject the model with foundational, multimodal medical knowledge from our rich MedAD-38K benchmark, and (ii) to align the model's output with our desired structured reasoning format, \texttt{<think>...</think><answer>...</answer>}. SFT is ideally suited for this purpose, as it efficiently enables the model to internalize the high-quality, manually verified data we have prepared.

During this stage, we maximize the likelihood of the ground-truth output $Y^*=(T^*,A^*)$ over the image-disjoint SFT subset $\mathcal{D}_{\mathrm{SFT}}$; the SFT/RL split is detailed in supplementary Section~\ref{app:data_split}. The negative log-likelihood objective is
\begin{equation}
    \mathcal{L}_{\text{SFT}}(\theta) = - \sum_{(I, Q, Y^*) \in \mathcal{D}_{\mathrm{SFT}}} \log \pi_\theta(Y^* | I, Q)
    \label{eq:sft_loss}
\end{equation}
This process fine-tunes the general-purpose LMM to grasp the nuances of medical imagery and terminology, while conditioning it to adopt the ``think-then-answer'' pattern. The resulting model $\pi_{\text{SFT}}$ serves as both a capable baseline and the starting point for the subsequent RL phase.

\subsection{Reasoning Reinforcement via Con-GRPO}
\label{sec:rl}

SFT alone only teaches the model to imitate the surface form of expert reasoning. It optimizes the likelihood of the reference output token by token, but it never verifies whether the thought $T$ actually entails the answer $A$. As a result, the model can still produce fluent reasoning that is not grounded in the image, or that quietly contradicts its own conclusion~\cite{gudibande2023false}. In clinical decision support, this disconnect is not a cosmetic flaw. A correct answer supported by unfaithful reasoning is as untrustworthy as a wrong one. We therefore add a reinforcement learning stage that rewards the model not merely for being right, but for being right through evidence-grounded reasoning.

To keep this stage lightweight, we build on Group Relative Policy Optimization (GRPO)~\cite{shao2024deepseekmath}. PPO~\cite{schulman2017proximal} relies on a separate value network of comparable size for advantage estimation, which is costly for LMMs. GRPO instead uses the mean reward of a sampled group as a dynamic baseline, and thus avoids this memory overhead. Con-GRPO inherits this efficiency and adds a reward designed to close the reasoning--answer gap. We fine-tune $\pi_{\text{SFT}}$ on the disjoint RL subset $\mathcal{D}_{\mathrm{RL}}$ to maximize
\begin{equation}
    J(\theta) = \mathbb{E}_{(I,Q) \sim \mathcal{D}_{\mathrm{RL}},\, Y \sim \pi_\theta(\cdot | I, Q)} [R(Y)]
    \label{eq:rl_objective}
\end{equation}

\subsubsection{Reward Function Design}
A high-quality diagnostic output should be well-formed, correct, and supported by reasoning that is faithful to the visual evidence. These three requirements are complementary and jointly capture the quality of the generated response. A deficiency in any component reduces the reliability of the generation. We therefore compose the reward from three corresponding terms and weight them equally:
\begin{equation}
    R(Y) = \lambda_{\text{fmt}} R_{\text{fmt}}(Y) + \lambda_{\text{acc}} R_{\text{acc}}(A, A^*) + \lambda_{\text{eac}} R_{\text{eac}},
    \label{eq:total_reward}
\end{equation}
where $\lambda_{\text{fmt}} = \lambda_{\text{acc}} = \lambda_{\text{eac}} = 1/3$. The ablation in Table~\ref{tab:reward_ablation} confirms that this balanced setting outperforms format-, accuracy-, or consistency-dominant alternatives.

The format reward $R_{\text{fmt}}$ is a binary signal that enforces the \texttt{<think>...</think><answer>...</answer>} structure. It returns $1$ when both tags are properly enclosed and $0$ otherwise. The accuracy reward scores the final prediction against the ground truth:
\begin{equation}
    R_{\text{acc}}(A, A^*) = \mathbb{I}(A = A^*).
    \label{eq:accuracy_reward}
\end{equation}

The third term, the Evidence-Aware Consistency Reward
$R_{\mathrm{eac}}$, is the core of Con-GRPO. A standard
consistency signal only examines whether the reasoning and the
answer agree. It does not determine whether the reasoning is
supported by the image and may therefore reward self-consistent
but hallucinated explanations. We instead introduce an
image-conditioned assessment using a domain-specific evaluator
$\mathcal{F}$, instantiated as MedGemma-4B~\cite{medgemma}.
We select a medical LMM because general-purpose models may
misinterpret complex medical images, which is one of the
limitations addressed in this work.

Given the tuple $(I,Q,A,T)$, the evaluator assigns five binary
judgments:
\emph{visual grounding}, $d_{\mathrm{vg}}$, which indicates
whether the evidence described in $T$ is supported by $I$;
\emph{task relevance}, $d_{\mathrm{tr}}$, which indicates
whether $T$ addresses the goal of $Q$;
\emph{answer support}, $d_{\mathrm{as}}$, which indicates
whether $T$ logically supports $A$;
\emph{no hallucination}, $d_{\mathrm{nh}}$, which indicates
the absence of fabricated findings; and
\emph{no contradiction}, $d_{\mathrm{nc}}$, which indicates
the absence of internal contradictions or conflicts with the
answer. Each judgment belongs to $\{0,1\}$, where $1$ denotes
the desirable outcome. Their mean defines the evidence-aware
consistency score:
\begin{equation}
S_{\mathrm{eac}}(I,Q,A,T)
=
\frac{
d_{\mathrm{vg}}
+
d_{\mathrm{tr}}
+
d_{\mathrm{as}}
+
d_{\mathrm{nh}}
+
d_{\mathrm{nc}}
}{5}.
\label{eq:eac_score}
\end{equation}

We condition this score on answer correctness. Reasoning attached
to an incorrect answer should not receive a positive consistency
reward, even when it appears coherent. Using the accuracy reward
defined in Equation~\ref{eq:accuracy_reward}, the Evidence-Aware
Consistency Reward is
\begin{equation}
R_{\mathrm{eac}}
=
R_{\mathrm{acc}}
\cdot
S_{\mathrm{eac}}(I,Q,A,T).
\label{eq:eac_reward}
\end{equation}

This gate establishes a clear division of responsibility between
the two rewards. The accuracy reward $R_{\mathrm{acc}}$
encourages the policy to produce the correct answer, whereas
$R_{\mathrm{eac}}$ distinguishes the quality of reasoning among
responses with correct answers. It therefore assigns lower rewards
to cases in which the model reaches the correct answer through
unsupported, hallucinated, or contradictory reasoning. Because
$\mathcal{F}$ may still produce imperfect judgments, we calibrate
it against human annotations before training. The complete
calibration protocol and agreement results are reported in
supplementary Section~\ref{app:judge_calibration}.

\subsubsection{Policy Optimization with Con-GRPO}
Following GRPO, for each input $(I,Q)$ the old policy $\pi_{\theta_{\mathrm{old}}}$ samples $G$ outputs $\{Y_i\}_{i=1}^{G}$. Their rewards are normalized within the group to obtain the completion-level advantage
\begin{equation}
    \hat A_i=\frac{R(Y_i)-\bar R}{\sigma_R+\delta},\quad
    \bar R=\frac{1}{G}\sum_{j=1}^{G}R(Y_j),
    \label{eq:advantage}
\end{equation}
where $\sigma_R$ is the group reward standard deviation, $\delta$ ensures numerical stability, and $\hat A_i$ is broadcast to all tokens of $Y_i$. Let
\begin{equation}
    r_{i,t}(\theta)=\frac{\pi_\theta(y_{i,t}\mid I,Q,y_{i,<t})}
    {\pi_{\theta_{\mathrm{old}}}(y_{i,t}\mid I,Q,y_{i,<t})}.
    \label{eq:policy_ratio}
\end{equation}
The clipped objective, with the frozen SFT model $\pi_{\mathrm{ref}}=\pi_{\mathrm{SFT}}$ used only as the KL reference, is
{\small
\begin{equation}
\label{eq:grpo_objective}
\begin{aligned}
J_{\mathrm{Con\text{-}GRPO}}(\theta)=\mathbb{E}\!\left[\frac{1}{G}\sum_{i=1}^{G}\frac{1}{|Y_i|}\sum_t
\left\{\min\!\left(
r_{i,t}\hat A_i,\right.\right.\right.\\[-2pt]
\left.\left.\left.
\operatorname{clip}(r_{i,t},1-\epsilon,1+\epsilon)\hat A_i
\right)-\beta D_{\mathrm{KL}}^{i,t}(\pi_\theta\|\pi_{\mathrm{ref}})\right\}\right],
\end{aligned}
\end{equation}
}
where $\epsilon$ clips the policy update and $\beta$ controls the KL penalty. The group-relative baseline avoids a learned value model, while the evidence-aware reward encourages visually grounded and answer-consistent reasoning.

\section{Experiments}
\label{sec:experiments}

In this section, we rigorously evaluate our proposed model, MedAD-R1. We aim to answer four questions: (1) Does MedAD-R1 outperform SOTA medical and general-purpose LMMs on our challenging benchmark? (2) Does its advantage persist on external datasets the model has never seen? (3) Does it genuinely reason more consistently, rather than merely answer more accurately? (4) Is the proposed framework tied to one particular backbone?

\subsection{Experimental Setup}
\label{sec:setup}

\begin{table*}[t]
    \centering
    \small
    \begin{tabular}{lcccccccccc}
        \toprule
        \textbf{Model} & \textbf{Params} & \textbf{Anat.} & \textbf{Anom.} & \textbf{Loc.} & \textbf{Mod.} & \textbf{Path.} & \textbf{Overall} & \textbf{Consist.} & \textbf{EOverall} & \textbf{EConsist.} \\
        \midrule
        Qwen3.5-0.8B & 0.8B & 94.22 & 57.88 & 33.74 & 93.45 & 14.93 & 76.43 & 92.12 & 68.12 & 88.75 \\
        InternVL3.5-1B & 1B & 95.04 & 50.09 & 33.53 & 89.27 & 15.52 & 73.10 & 93.53 & 69.81 & 91.53 \\
        Qwen3VL-2B & 2B & 91.20 & 55.54 & 37.08 & 92.57 & 10.75 & 74.91 & 94.84 & 67.35 & 94.41 \\
        Qwen2.5VL-3B & 3B & 78.56 & 50.29 & 30.63 & 78.97 & 11.04 & 64.90 & 93.25 & 51.07 & 92.01 \\
        MiMo-VL-7B & 7B & 61.71 & 55.20 & 30.04 & 63.71 & 28.81 & 56.86 & 94.57 & 39.49 & 93.78 \\
        Qwen2.5VL-7B & 7B & 75.43 & 56.10 & 26.90 & 82.28 & 15.52 & 66.30 & 94.17 & 50.12 & 91.84 \\
        MiniCPM-V-4.5 & 9B & 90.01 & 54.65 & 26.64 & 93.24 & 6.04 & 73.37 & 94.96 & 59.33 & 95.10 \\
        Qwen3.5-9B & 9B & 94.39 & 61.27 & 34.40 & 96.11 & 20.90 & 78.38 & 94.25 & 75.97 & 95.05 \\
        GLM-4.6V-Flash & 9B & 87.55 & 58.04 & 48.21 & 87.11 & 16.12 & 74.15 & 93.43 & 66.80 & 94.43 \\
        Llama-3.2-Vision & 11B & 92.18 & 55.65 & 44.35 & 92.88 & 15.97 & 76.12 & 94.04 & 68.70 & 95.67 \\
        Gemma4-26B & 26B & 93.80 & 65.78 & 36.58 & 94.28 & 12.76 & 79.18 & \underline{96.95} & 76.50 & \underline{96.01} \\
        Qwen3.6-27B & 27B & 93.82 & 60.16 & 39.71 & 90.67 & 8.36 & 76.76 & 92.22 & 77.26 & 88.93 \\
        Qwen3VL-32B & 32B & 84.62 & 53.48 & 37.83 & 90.44 & 18.43 & 71.84 & 96.23 & 68.73 & 95.66 \\
        InternVL3.5-38B & 38B & 94.13 & 62.06 & 50.52 & 94.67 & 17.80 & 79.79 & 94.00 & 79.72 & 94.49 \\
        \midrule
        MedVLM-R1$^{*}$ & 2B & 81.80 & 51.07 & 20.46 & 81.51 & 6.27 & 65.74 & 38.90 & 54.57 & 36.85 \\
        HuatuoGPT-Vision$^{*}$ & 7B & 84.09 & 45.46 & 34.08 & 85.52 & 4.70 & 67.38 & 32.50 & 62.90 & 29.25 \\
        Lingshu-7B$^{*}$ & 7B & 80.44 & 53.03 & 29.31 & 76.52 & 12.84 & 65.41 & 95.60 & 54.11 & 94.03 \\
        LLaVA-Med$^{*}$ & 7B & 76.85 & 50.10 & 22.35 & 74.90 & 7.16 & 62.24 & 31.50 & 55.25 & 28.40 \\
        \midrule
        InternVL3.5-8B$^{\dagger}$ & 8B & 95.77 & 76.66 & 83.18 & \underline{96.16} & 27.27 & 88.53 & 95.39 & 83.55 & 94.67 \\
        Qwen3VL-8B$^{\dagger}$ & 8B & 95.87 & 79.25 & 70.11 & 95.53 & 72.46 & 88.01 & 93.33 & 75.48 & 91.92 \\
        Qwen3.6-27B$^{\dagger}$ & 27B & \underline{95.91} & \underline{87.38} & \underline{91.44} & 95.61 & \underline{82.11} & \underline{92.75} & 92.10 & \underline{84.27} & 90.94 \\
        \midrule
        \textbf{MedAD-R1 (Ours)} & 0.8B & \textbf{98.96} & \textbf{88.30} & \textbf{91.49} & \textbf{99.57} & \textbf{85.57} & \textbf{95.12} & \textbf{97.26} & \textbf{90.09} & \textbf{96.26} \\
        \bottomrule
    \end{tabular}
    \caption{Comparison on MedAD-38K and the external zero-shot test set. Anat.: anatomy identification; Anom.: anomaly detection; Loc.: lesion localization; Mod.: modality classification; Path.: pathology characterization. Overall and Consist. denote accuracy and reasoning--answer consistency on MedAD-38K; EOverall and EConsist. denote the corresponding external-set metrics. Overall values are micro-averaged over applicable VQA instances. $^{*}$ marks medical LMMs and $^{\dagger}$ marks larger models fine-tuned on our training data. Boldface denotes the best result and underlining the second best. Means and standard deviations over five inference passes are reported in supplementary Tables~\ref{tab:indomain_accuracy_detailed} and~\ref{tab:external_results}.}
    \label{tab:main_results}
\end{table*}

\begin{table}[t]
    \centering
    \small
    \setlength{\tabcolsep}{3pt}
    \begin{tabular}{llccc}
        \toprule
        \textbf{Backbone} & \textbf{Setting} & \textbf{Overall$\uparrow$} & \textbf{Consist.$\uparrow$} & \textbf{EOverall$\uparrow$} \\
        \midrule
        \multirow{3}{*}{Qwen3.5-0.8B}
            & Zero-shot   & 76.43 & 92.12 & 68.12 \\
            & SFT only    & 90.89 & 94.35 & 88.36 \\
            & Con-GRPO    & 95.12 & 97.26 & 90.09 \\
        \midrule
        \multirow{3}{*}{InternVL3.5-1B}
            & Zero-shot   & 73.10 & 93.53 & 69.81 \\
            & SFT only    & 86.40 & 93.75 & 81.63 \\
            & Con-GRPO    & 92.31 & 96.73 & 83.72 \\
        \midrule
        \multirow{3}{*}{Qwen3VL-2B}
            & Zero-shot   & 74.91 & 94.84 & 67.35 \\
            & SFT only    & 86.72 & 92.84 & 86.00 \\
            & Con-GRPO    & 90.83 & 96.84 & 87.65 \\
        \midrule
        \multirow{3}{*}{Qwen2.5VL-3B}
            & Zero-shot   & 64.90 & 93.25 & 51.07 \\
            & SFT only    & 87.97 & 92.84 & 85.62 \\
            & Con-GRPO    & 92.85 & 96.15 & 85.44 \\
        \bottomrule
    \end{tabular}
    \caption{Cross-backbone effectiveness and ablation. For each backbone we report zero-shot, SFT-only, and full Con-GRPO. Overall and Consist.\ are measured on MedAD-38K; EOverall is measured on the external set. All metrics are reported as percentages, and higher is better ($\uparrow$).}
    \label{tab:backbone_ablation}
\end{table}

\paragraph{Datasets.}
We use the image-disjoint MedAD-38K test split and five source-disjoint external datasets: ColonDB~\cite{tajbakhsh2015automated}, HeadCT~\cite{salehi2021multiresolution}, EndoTect~\cite{hicks2021endotect}, TN3K~\cite{gong2021multi}, and NMSC~\cite{thomas2021non}. Their VQA pairs follow the same task formats but require no reference CoT. Split and dataset details are provided in supplementary Sections~\ref{app:data_split} and~\ref{app:external_data}.

\paragraph{Evaluation Metrics.}
We report per-task and Overall Accuracy on the test set. Overall Accuracy is micro-averaged over all applicable VQA instances and is not the arithmetic mean of the five task-level accuracies. Since accuracy alone cannot reveal whether a correct answer rests on sound reasoning, we additionally measure reasoning consistency through a back-inference protocol: a panel of three multimodal judges receives the question, answer options, and generated reasoning, but neither the medical image nor the model's own answer, and each infers which option the reasoning supports. A sample is consistent when the majority-voted inference matches the model's answer. If all three judges infer different options, the panel verdict is treated as \texttt{INSUFFICIENT}. The panel shares no model with the training-time evaluator, so the measurement is not circular. Judge separation and the full protocol are reported in supplementary Sections~\ref{app:judge_separation} and~\ref{app:consistency_eval}; agreement and in-domain/external dimension results appear in Tables~\ref{tab:judge_agreement},~\ref{tab:consistency_test_results}, and~\ref{tab:consistency_etenl_results}.

\paragraph{Implementation Details.}
MedAD-R1 uses Qwen3.5-0.8B~\cite{qwen35blog} as its primary backbone, trained with a LoRA-based SFT stage followed by Con-GRPO. To verify that the framework is not tied to this backbone, we repeat the same two-stage protocol on Qwen3VL-2B, Qwen2.5VL-3B, and InternVL3.5-1B under an identical data split and optimization setup. Full training and deployment settings are reported in supplementary Section~\ref{app:implementation}, and the controlled inference-efficiency evaluation is reported in Section~\ref{app:efficiency}.

\paragraph{Baselines.}
We compare against three groups. (i) \textbf{Domain-specific medical LMMs} ($^{*}$): HuatuoGPT-Vision~\cite{huatuogpt}, Lingshu~\cite{lingshu}, LLaVA-Med~\cite{llava-med}, and MedVLM-R1~\cite{medvlm}. (ii) \textbf{General-purpose LMMs} evaluated zero-shot, spanning 0.8B to 38B: the Qwen2.5VL~\cite{qwen2.5vl}, Qwen3~\cite{bai2025qwen3}, and InternVL3.5~\cite{internvl3} families, together with Gemma4-26B~\cite{team2026gemma}, GLM-4.6V-Flash~\cite{hong2025glm}, MiMo-VL-7B~\cite{mimovl}, MiniCPM-V-4.5-9B~\cite{minicpm}, and Llama-3.2-11B-Vision~\cite{llama3}. (iii) \textbf{Larger models fine-tuned on our data} ($^{\dagger}$): Qwen3VL-8B, InternVL3.5-8B, and Qwen3.6-27B~\cite{qwen36_27b}, trained on the same split under our Stage-1 recipe. This third group isolates the contribution of our framework from that of in-domain data, since a zero-shot comparison alone cannot separate the two. Each fixed checkpoint is evaluated through five independent stochastic vLLM inference passes without a fixed inference seed; prompts, option order, images, and maximum output length remain unchanged. Accuracy is reported as the mean and standard deviation, with no retraining between passes; exact sampling parameters are given in supplementary Section~\ref{app:implementation}. Consistency is evaluated once by the three-judge panel over the pooled generations.

\subsection{Comparison with SOTA Models}

\paragraph{In-domain (Q1).}
With only 0.8B parameters, MedAD-R1 reaches 95.12\% Overall Accuracy on MedAD-38K and ranks first on every task (Table~\ref{tab:main_results}), surpassing all zero-shot baselines including much larger ones. Even after in-domain SFT, the strongest such baseline, Qwen3.6-27B, reaches 92.75\%, so a model 34 times larger trained on the same data still trails ours by 2.37 points. The gain therefore stems from the framework rather than from in-domain data alone.

\paragraph{External generalization (Q2).}
The advantage persists on the five source-disjoint external datasets, where MedAD-R1 attains the best external accuracy of 90.09\%. The margin over the fine-tuned baselines widens here: Qwen3.6-27B$^{\dagger}$ drops to 84.27\%, 5.82 points below ours. The lightweight model thus generalizes to unseen imaging sources more reliably than larger models trained on identical data.

\paragraph{Reasoning consistency (Q3).}
MedAD-R1 achieves the highest consistency rates both in-domain (97.26\%) and externally (96.26\%). Although Qwen3.6-27B$^{\dagger}$ approaches its accuracy, its consistency is lower at 92.10\%, showing that correct answers do not guarantee consistent reasoning. Medical LMMs perform substantially worse, with LLaVA-Med reaching only 31.50\%; examples appear in supplementary Section~\ref{app:qualitative_success}.

\paragraph{Backbone independence (Q4).}
Con-GRPO improves every backbone over its SFT-only counterpart by 4.1 to 5.9 points in Overall Accuracy, with parallel gains in consistency (Table~\ref{tab:backbone_ablation}), supporting its generality across families and scales; per-task results are in supplementary Table~\ref{tab:cross_backbone}.

\subsection{Ablation Study}

We conduct two ablations on the Qwen3.5-0.8B backbone (Tables~\ref{tab:ablation_results} and~\ref{tab:reward_ablation}). Since a central claim is that the Evidence-Aware Consistency Reward improves reasoning rather than the answer alone, we report consistency alongside accuracy.

The two-stage design is essential. Without SFT, the RL-only model reaches only 81.84\%, confirming that reinforcement learning from an unaligned policy is ineffective, whereas SFT alone already provides a strong 90.89\% starting point. Adding a conventional accuracy reward (SFT + Acc) raises accuracy to 92.76\% but does not improve consistency, which decreases slightly from 94.35\% to 93.71\%, since rewarding only the final answer permits correct predictions reached through unsupported reasoning. The Evidence-Aware Consistency Reward lifts both accuracy (95.12\%) and, more importantly, consistency (97.26\%): the reward improves not merely whether the answer is correct, but whether the reasoning behind it is faithful and coherent. Finally, the balanced weights outperform format-, accuracy-, or consistency-dominant settings, confirming that the three terms are complementary.

\begin{table}[tb]
    \centering
    \small
    \begin{tabular}{lcc}
        \toprule
        Configuration & Overall$\uparrow$ & Consist.$\uparrow$ \\
        \midrule
        RL-only (no SFT)         & 81.84 $\pm$ 2.41 & 88.90 \\
        SFT only                 & 90.89 $\pm$ 0.27 & 94.35 \\
        SFT + Acc                & 92.76 $\pm$ 0.44 & 93.71 \\
        \textbf{SFT + Acc + EAC (Ours)} & \textbf{95.12 $\pm$ 0.04} & \textbf{97.26} \\
        \bottomrule
    \end{tabular}
    \caption{Ablation of the training stages and reward terms on MedAD-38K.
    Acc: accuracy reward only. EAC: Evidence-Aware Consistency Reward.
    All values are percentages; higher is better. }
    \label{tab:ablation_results}
\end{table}

\begin{table}[tb]
    \centering
    \small
    \begin{tabular}{lcccc}
        \toprule
        Configuration & $\lambda_{\text{fmt}}$ & $\lambda_{\text{acc}}$ & $\lambda_{\text{eac}}$ & Overall$\uparrow$ \\
        \midrule
        Format-focused      & 0.8 & 0.1 & 0.1 & 89.34 \\
        Accuracy-focused    & 0.1 & 0.8 & 0.1 & 93.12 \\
        Consistency-focused & 0.1 & 0.1 & 0.8 & 93.88 \\
        \textbf{Balanced (Ours)} & \textbf{1/3} & \textbf{1/3} & \textbf{1/3} & \textbf{95.12} \\
        \bottomrule
    \end{tabular}
    \caption{Ablation of Con-GRPO reward weights on MedAD-38K.
    $\lambda_{\text{eac}}$ weights the Evidence-Aware Consistency Reward.
    }
    \label{tab:reward_ablation}
\end{table}


\section{Conclusion}
We addressed two obstacles to reasoning-capable MedAD: fragmented data and the tendency of standard SFT to leave the reasoning process unconstrained. We introduced MedAD-38K, a large-scale, multimodal and multicenter benchmark with quality-controlled CoT annotations, together with a unified zero-shot protocol over five external datasets. On this foundation we proposed a two-stage framework whose core is Con-GRPO with an Evidence-Aware Consistency Reward, scoring the reasoning for visual grounding, task relevance, answer support, and the absence of hallucination and contradiction. The resulting MedAD-R1 reaches state-of-the-art accuracy and reasoning consistency on both MedAD-38K and the external datasets, and these gains transfer across four backbones, while the primary MedAD-R1 model contains only 0.8B parameters. We hope MedAD-38K and Con-GRPO help advance trustworthy and interpretable medical anomaly detection, and we plan to extend the framework to open-ended diagnostic reasoning.

\bibliography{references}

\newpage
\input{supplementary_material}


\end{document}

%% file: supplementary_material.tex

\onecolumn
\raggedbottom
\setcounter{topnumber}{5}
\setcounter{bottomnumber}{3}
\setcounter{totalnumber}{8}
\renewcommand{\topfraction}{0.92}
\renewcommand{\bottomfraction}{0.85}
\renewcommand{\textfraction}{0.08}
\renewcommand{\floatpagefraction}{0.82}
\setlength{\floatsep}{10pt plus 2pt minus 2pt}
\setlength{\textfloatsep}{12pt plus 2pt minus 3pt}
\setlength{\intextsep}{10pt plus 2pt minus 2pt}

\newenvironment{promptblock}{%
    \par\nopagebreak[4]\medskip\noindent\hspace{0.02\textwidth}%
    \begin{minipage}{0.96\textwidth}%
    \hrule height 0.5pt\vspace{3pt}
}{%
    \vspace{2pt}\hrule height 0.5pt
    \end{minipage}\par\medskip
}

\appendix
\setcounter{secnumdepth}{2}

\setcounter{section}{0}
\renewcommand{\thesection}{\Alph{section}}
\renewcommand{\thesubsection}{\thesection.\arabic{subsection}}
\renewcommand{\thetable}{\Alph{section}.\arabic{table}}
\renewcommand{\thefigure}{\Alph{section}.\arabic{figure}}
\renewcommand{\theequation}{\Alph{section}.\arabic{equation}}

\makeatletter
\renewcommand\section{\@startsection{section}{1}{\z@}%
    {-2.0ex plus -0.5ex minus -.2ex}%
    {3pt plus 2pt minus 1pt}%
    {\Large\bfseries\raggedright}}
\renewcommand\subsubsection{\@startsection{subsubsection}{3}{\z@}%
    {-1.5ex plus -0.5ex minus -.2ex}%
    {2pt plus 1pt minus 1pt}%
    {\normalsize\bfseries\raggedright}}
\@addtoreset{table}{section}
\@addtoreset{figure}{section}
\@addtoreset{equation}{section}
\setlength{\@fptop}{0pt}
\setlength{\@fpbot}{0pt plus 1fil}
\makeatother

\section*{Appendices}

Our appendices include the following sections:
\begin{itemize}\itemsep0pt\parsep0pt
    \item \textbf{Sec.~\ref{app:dataset}: Details of the MedAD-38K Dataset.}
    \begin{itemize}\itemsep0pt\parsep0pt
        \item Dataset Composition and Statistics.
        \item VQA Construction and Question Templates.
        \item Five-Region Lesion Localization.
        \item CoT Generation, Annotation Statistics, and Manual Quality Control.
        \item Data Split and Leakage Prevention.
    \end{itemize}
    \item \textbf{Sec.~\ref{app:evaluators}: Reliability of the Language-Model Evaluators.}
    \begin{itemize}\itemsep0pt\parsep0pt
        \item Calibration of the Reward Evaluator.
        \item Separation Between Training and Evaluation Judges.
        \item Consistency Evaluation Protocol.
    \end{itemize}
    \item \textbf{Sec.~\ref{app:extended}: Extended Experimental Results.}
    \begin{itemize}\itemsep0pt\parsep0pt
        \item Detailed Reasoning-Quality Results.
        \item External-Dataset Results.
        \item Cross-Backbone Results.
        \item Implementation Details.
    \end{itemize}
    \item \textbf{Sec.~\ref{app:efficiency}: Inference Efficiency and Deployment.}
    \begin{itemize}\itemsep0pt\parsep0pt
        \item Measurement Protocol.
        \item Natural-Generation and Length-Controlled Results.
    \end{itemize}
    \item \textbf{Sec.~\ref{app:qualitative}: Qualitative Comparisons.}
    \begin{itemize}\itemsep0pt\parsep0pt
        \item Cases Where MedAD-R1 Improves Over the Baselines.
    \end{itemize}
\end{itemize}

\section{Details of the MedAD-38K Dataset}
\label{app:dataset}

\subsection{Dataset Composition and Statistics}
\label{app:composition}

MedAD-38K contains 38,422 medical images and 130,146 VQA instances collected
from 16 public datasets. The name MedAD-38K refers to the number of images
rather than the number of VQA instances. Each image is associated with an
average of 3.39 questions, depending on the annotations available in its source
dataset. The benchmark covers 10 imaging modalities and 10 anatomical regions.

Table~\ref{tab:source_statistics} reports the source-level composition. The
benchmark contains 19,617 normal images and 18,805 abnormal images,
corresponding to 51.1\% and 48.9\% of the complete image collection.

\begin{table}[htbp]
    \centering
    \setlength{\tabcolsep}{6pt}
    \renewcommand{\arraystretch}{1.1}
    \begin{tabular}{lllrrrr}
        \toprule
        \textbf{Source} & \textbf{Modality} & \textbf{Anatomy} &
        \textbf{Normal} & \textbf{Abnormal} & \textbf{Images} &
        \textbf{VQA Instances} \\
        \midrule
        ATLAS        & CE-MRI     & Liver            & 2,571 & 2,173 & 4,744  & 16,405 \\
        BACH         & Microscopy & Breast           & 100   & 300   & 400    & 1,500  \\
        BUSI         & Ultrasound & Breast           & 133   & 647   & 780    & 2,987  \\
        Br35H        & MRI        & Brain            & 1,500 & 801   & 2,301  & 7,704  \\
        BraTS2021    & MRI        & Brain            & 8,179 & 3,109 & 11,288 & 36,973 \\
        CVC-ClinicDB & Endoscopy  & Gastrointestinal & 0     & 612   & 612    & 2,448  \\
        Chest        & X-ray      & Chest            & 1,583 & 4,273 & 5,856  & 17,568 \\
        DDTI         & Ultrasound & Thyroid          & 0     & 637   & 637    & 2,548  \\
        IDRiD        & Fundus     & Retina           & 168   & 348   & 516    & 2,244  \\
        ISIC2018     & Dermoscopy & Skin             & 0     & 2,594 & 2,594  & 10,376 \\
        Kvasir-SEG   & Endoscopy  & Gastrointestinal & 0     & 1,000 & 1,000  & 4,000  \\
        MLS          & Microscopy & Lymph-Node       & 0     & 374   & 374    & 1,496  \\
        MosMed       & CT         & Lung             & 0     & 785   & 785    & 3,140  \\
        PH2          & Dermoscopy & Skin             & 0     & 200   & 200    & 800    \\
        PediDemi     & MRI        & Brain            & 0     & 118   & 118    & 472    \\
        RESC         & OCT        & Retina           & 5,383 & 834   & 6,217  & 19,485 \\
        \midrule
        \textbf{Total} & & & \textbf{19,617} & \textbf{18,805} &
        \textbf{38,422} & \textbf{130,146} \\
        \bottomrule
    \end{tabular}
    \caption{Source-level composition of MedAD-38K. VQA counts differ across
    sources because lesion localization and pathology characterization are
    generated only when the required source annotations are available.}
    \label{tab:source_statistics}
\end{table}

Tables~\ref{tab:modality_statistics} and~\ref{tab:anatomy_statistics} report the
modality and anatomy distributions.

\begin{table}[htbp]
    \centering
    \begin{minipage}[t]{0.47\textwidth}
        \centering
        \setlength{\tabcolsep}{10pt}
        \renewcommand{\arraystretch}{1.1}
        \begin{tabular}{lrr}
            \toprule
            \textbf{Modality} & \textbf{Images} & \textbf{Proportion} \\
            \midrule
            MRI        & 13,707 & 35.7\% \\
            OCT        & 6,217  & 16.2\% \\
            X-ray      & 5,856  & 15.2\% \\
            CE-MRI     & 4,744  & 12.3\% \\
            Dermoscopy & 2,794  & 7.3\%  \\
            Endoscopy  & 1,612  & 4.2\%  \\
            Ultrasound & 1,417  & 3.7\%  \\
            CT         & 785    & 2.0\%  \\
            Microscopy & 774    & 2.0\%  \\
            Fundus     & 516    & 1.3\%  \\
            \bottomrule
        \end{tabular}
        \caption{Image distribution across the 10 imaging modalities.}
        \label{tab:modality_statistics}
    \end{minipage}
    \hfill
    \begin{minipage}[t]{0.47\textwidth}
        \centering
        \setlength{\tabcolsep}{10pt}
        \renewcommand{\arraystretch}{1.1}
        \begin{tabular}{lrr}
            \toprule
            \textbf{Anatomy} & \textbf{Images} & \textbf{Proportion} \\
            \midrule
            Brain            & 13,707 & 35.7\% \\
            Retina           & 6,733  & 17.5\% \\
            Chest            & 5,856  & 15.2\% \\
            Liver            & 4,744  & 12.3\% \\
            Skin             & 2,794  & 7.3\%  \\
            Gastrointestinal & 1,612  & 4.2\%  \\
            Breast           & 1,180  & 3.1\%  \\
            Lung             & 785    & 2.0\%  \\
            Thyroid          & 637    & 1.7\%  \\
            Lymph-Node       & 374    & 1.0\%  \\
            \bottomrule
        \end{tabular}
        \caption{Image distribution across the 10 anatomical regions.}
        \label{tab:anatomy_statistics}
    \end{minipage}
\end{table}

Table~\ref{tab:task_statistics} summarizes the task-level distribution. Modality
classification, anatomy identification, and anomaly detection are available for
every image. Lesion localization is generated only for abnormal images with
valid segmentation masks. Pathology characterization is generated only when the
source dataset provides an explicit disease label.

\begin{table}[htbp]
    \centering
    \begin{minipage}[t]{0.47\textwidth}
        \centering
        \setlength{\tabcolsep}{10pt}
        \renewcommand{\arraystretch}{1.1}
        \begin{tabular}{lr}
            \toprule
            \textbf{Task} & \textbf{VQA Instances} \\
            \midrule
            Modality Classification    & 38,422 \\
            Anatomy Identification     & 38,422 \\
            Anomaly Detection          & 38,422 \\
            Lesion Localization        & 13,858 \\
            Pathology Characterization & 1,022  \\
            \midrule
            \textbf{Total}             & \textbf{130,146} \\
            \bottomrule
        \end{tabular}
        \caption{Numbers of VQA instances for the five MedAD tasks.}
        \label{tab:task_statistics}
    \end{minipage}
    \hfill
    \begin{minipage}[t]{0.47\textwidth}
        \centering
        \setlength{\tabcolsep}{10pt}
        \renewcommand{\arraystretch}{1.1}
        \begin{tabular}{lr}
            \toprule
            \textbf{Pathology} & \textbf{Images} \\
            \midrule
            Follicular Lymphoma          & 139 \\
            Mantle Cell Lymphoma         & 122 \\
            Chronic Lymphocytic Leukemia & 113 \\
            Benign                       & 100 \\
            In Situ Carcinoma            & 100 \\
            Invasive Carcinoma           & 100 \\
            IDRiD grading labels (aggregated) & 348 \\
            \midrule
            \textbf{Total}               & \textbf{1,022} \\
            \bottomrule
        \end{tabular}
        \caption{Distribution of source-provided pathology labels. IDRiD
        grading labels are reported only in aggregate.}
        \label{tab:pathology_statistics}
    \end{minipage}
\end{table}

\subsection{VQA Construction}
\label{app:vqa_construction}

Each VQA instance is constructed from the labels, masks, or metadata provided by
its source dataset. Language models are not used to determine the correct
answers. Table~\ref{tab:task_mapping} summarizes the supervision source and the
candidate-answer pool for each task.

\begin{table}[htbp]
    \centering
    \setlength{\tabcolsep}{8pt}
    \renewcommand{\arraystretch}{1.1}
    \begin{tabular}{llll}
        \toprule
        \textbf{Task} & \textbf{Source of Correct Answer} &
        \textbf{Candidate Pool} & \textbf{Applicability} \\
        \midrule
        Modality Classification    & Source modality metadata  & 10 modalities            & All images \\
        Anatomy Identification     & Source anatomy metadata   & 10 anatomical regions    & All images \\
        Anomaly Detection          & Source normality label    & Yes / No                 & All images \\
        Lesion Localization        & Source segmentation mask  & Five spatial regions     & Abnormal images with masks \\
        Pathology Characterization & Source disease label      & Predefined pathology set & Images with disease labels \\
        \bottomrule
    \end{tabular}
    \caption{Mapping from source annotations to the five VQA tasks.}
    \label{tab:task_mapping}
\end{table}

For each task, we prepare ten semantically equivalent question templates to
reduce sensitivity to wording. One template is sampled when constructing each
VQA instance. Answer options are then randomized to reduce positional shortcuts.
Anomaly detection uses the two options Yes and No. For the other tasks, three
incorrect options are sampled from the corresponding candidate pool, combined
with the correct answer, and randomly shuffled. The shuffled choices are
assigned option labels from A to D.

Some anomaly-detection templates use reversed polarity, such as asking whether
an image is unremarkable or within normal limits. For these templates, the Yes
or No answer is deterministically mapped according to the semantic polarity of
the question. This prevents the same source label from being assigned
mechanically without considering the wording of the selected template.

Lesion localization is generated only for abnormal images with valid
segmentation masks. Pathology characterization is generated only for BACH, MLS,
and IDRiD images with source-provided disease labels. Because IDRiD provides
both pixel-level lesion masks and image-level grading labels, its abnormal
images contribute to both lesion localization and pathology characterization.
Tasks requiring unavailable annotations are omitted rather than completed using
model-generated labels.

Training instances use a structured conversational format. The user input
contains the image, the multiple-choice question, and an instruction requesting
the reasoning inside \texttt{<think></think>} tags and the selected option
inside \texttt{<answer></answer>} tags. The target response contains the
verified CoT and the correct option label. Test instances use the same
user-side format, while the reference CoT is not exposed to the evaluated model.

\subsubsection{Question Templates}

The complete question-template pools are listed below.

\paragraph{Modality Classification.}
\begin{enumerate}\itemsep0pt
    \item Based on its visual properties, what is the source modality of this image?
    \item Can you determine the technology used to generate this scan?
    \item Could you identify the type of scan shown?
    \item From which imaging technology does this scan originate?
    \item How would you classify the modality of this medical image?
    \item Is this scan a CT, an MRI, an X-ray, or another type of modality?
    \item What is the imaging modality being used in this picture?
    \item What kind of medical imaging are we looking at here?
    \item What method of image acquisition does this appear to be?
    \item Which imaging technique was employed to create this image?
\end{enumerate}

\paragraph{Anatomy Identification.}
\begin{enumerate}\itemsep0pt
    \item Could you name the main body part under observation?
    \item The image features which major anatomical structure?
    \item What anatomy is primarily shown in the image?
    \item What bodily region is visible here?
    \item What is the specific anatomical focus of this imaging study?
    \item What organ or structure is the subject of this examination?
    \item What part of the body is depicted in this scan?
    \item Which anatomical region is being displayed?
    \item Which area of the body was targeted for this scan?
    \item Which part of the anatomy are we currently examining?
\end{enumerate}

\paragraph{Anomaly Detection.}
\begin{enumerate}\itemsep0pt
    \item Are any pathologies visible in this scan?
    \item Are there any clinically significant findings visible?
    \item Can any abnormalities be detected upon inspection?
    \item Does the anatomy in this image appear to be unremarkable?
    \item Does the depicted tissue show any deviation from a healthy state?
    \item Does this scan reveal any signs of pathology?
    \item Is there an anomaly present in the image?
    \item Is there evidence of pathological change in this image?
    \item Is this scan considered to be within normal limits?
    \item Would you classify this scan as normal or abnormal?
\end{enumerate}

\paragraph{Lesion Localization.}
\begin{enumerate}\itemsep0pt
    \item A pathological finding is visible. Can you specify its location relative to nearby structures?
    \item An anomaly has been identified in this image. Where is it precisely located?
    \item In which anatomical quadrant or area does the confirmed finding reside?
    \item Given that a lesion is present, can you identify its region within the anatomy?
    \item Since a lesion has been detected, which region should receive the closest attention?
    \item The image contains a confirmed abnormality. How would you describe its spatial position?
    \item There is an area of concern in this medical image. Where is it located?
    \item This is an abnormal scan. Which part of the organ or tissue contains the pathology?
    \item This scan shows a pathology. What is its position relative to the image center?
    \item Given that this scan is abnormal, which spatial region contains most of the lesion?
\end{enumerate}

\paragraph{Pathology Characterization.}
\begin{enumerate}\itemsep0pt
    \item The image displays a known pathology. What is its specific type?
    \item Characterize the type of pathology observed.
    \item Describe the nature of the abnormality.
    \item What kind of lesion is this?
    \item Classify the specific pathology shown.
    \item What is the diagnosis based on the visual evidence?
    \item Identify the type of lesion present.
    \item Which pathology category best describes the observed lesion?
    \item What are the characteristics of this pathology?
    \item Provide a specific classification for the observed abnormality.
\end{enumerate}

\subsection{Five-Region Lesion Localization}
\label{app:five_region}

Lesion-localization labels are derived directly from segmentation masks provided
by the source datasets. Each image and its corresponding mask are represented on
the same square spatial grid with side length $L$.

The horizontal and vertical centerlines first divide the image into four
quadrants. A concentric square with side length $s$ is then defined as the
center region. Because this square is centered at the intersection of the two
centerlines, it overlaps each quadrant by an area of $s^{2}/4$. After removing
this overlap, the area of each directional region is
\begin{equation}
    A_{\mathrm{dir}} = \frac{L^{2}}{4} - \frac{s^{2}}{4}
    = \frac{L^{2}-s^{2}}{4},
\end{equation}
while the center region has area
\begin{equation}
    A_{\mathrm{center}} = s^{2}.
\end{equation}

To make the center region and the four directional regions equal in area, we
impose $A_{\mathrm{center}} = A_{\mathrm{dir}}$, which gives
\begin{equation}
    s^{2} = \frac{L^{2}-s^{2}}{4}.
\end{equation}
Rearranging yields
\begin{align}
    4s^{2} &= L^{2}-s^{2}, \\
    5s^{2} &= L^{2}, \\
    s &= \frac{L}{\sqrt{5}} .
\end{align}
The side-length ratio is therefore
\begin{equation}
    \frac{s}{L} = \frac{1}{\sqrt{5}} \approx 0.4472,
\end{equation}
and each of the five regions occupies exactly one fifth of the image:
\begin{equation}
    A_{\mathrm{center}} = s^{2} = \frac{L^{2}}{5},
    \qquad
    A_{\mathrm{dir}} = \frac{1}{4}\left(L^{2}-\frac{L^{2}}{5}\right)
    = \frac{L^{2}}{5}.
\end{equation}

The boundaries of the center square are
\begin{equation}
    x_{\mathrm{left}} = y_{\mathrm{top}} = \frac{L-s}{2}
    = \frac{L}{2}\left(1-\frac{1}{\sqrt{5}}\right),
    \qquad
    x_{\mathrm{right}} = y_{\mathrm{bottom}} = \frac{L+s}{2}
    = \frac{L}{2}\left(1+\frac{1}{\sqrt{5}}\right).
\end{equation}
In normalized image coordinates, the center region therefore spans
\begin{equation}
    \left[\frac{1}{2}\left(1-\frac{1}{\sqrt{5}}\right),\;
          \frac{1}{2}\left(1+\frac{1}{\sqrt{5}}\right)\right]
    \approx [0.2764,\,0.7236]
\end{equation}
along both dimensions.

Let $M(x,y)\in\{0,1\}$ denote the source-provided anomaly mask, and let
$\Omega_{k}$ denote one of the five disjoint spatial regions. The number of
anomalous pixels in region $k$ is
\begin{equation}
    n_{k} = \sum_{(x,y)\in\Omega_{k}} M(x,y),
\end{equation}
and the lesion-localization label is assigned by
\begin{equation}
    \hat{k} = \arg\max_{k} \, n_{k}.
\end{equation}

When a lesion crosses multiple regions or contains several disconnected
components, the complete anomaly mask is used, so the assigned label corresponds
to the region containing the largest total number of anomalous pixels. This
procedure does not rely on a bounding-box center or a mask centroid.

The construction introduces an explicit center category while avoiding the fine
spatial fragmentation of a $3\times3$ grid. Equalizing the five region areas
also prevents the center region from receiving a smaller prior area than the
directional regions.

\subsection{CoT Generation}
\label{app:cot_generation}

The CoT annotation process consists of auxiliary image-description generation,
task-specific rationale generation, and manual verification. The first two
stages are described here, and the verification protocol is reported in
Section~\ref{app:quality_control}.

\subsubsection{Auxiliary Image Descriptions}

MedGemma-27B generates an auxiliary medical description for each image. For an
abnormal sample, the model receives the original image, its source-provided
segmentation mask, and verified metadata including the imaging modality,
anatomical region, abnormality status, and lesion location. For a normal sample,
the model receives the image together with modality, anatomy, and normality
metadata. Providing these annotations is deliberate because this stage performs
supervised description synthesis rather than label-blind evaluation. The prompts
nevertheless require the model to distinguish visible observations from
annotation context and to flag insufficient visual support instead of inventing
evidence.

\paragraph{Prompt for abnormal samples.}\mbox{}\par
\begin{promptblock}
\begin{verbatim}
You are preparing an evidence-grounded description of a medical image.
You are given the image, its source-provided anomaly mask, and verified
annotation metadata.

Metadata:
  Modality: {modality}
  Anatomy: {anatomy}
  Status: abnormal
  Lesion region: {region}

Use the metadata as annotation context, not as a substitute for visual
evidence. Inspect the image first, then write a concise structured
description covering:
  1. Imaging characteristics of the modality.
  2. The visible anatomical structures.
  3. Visually discernible findings within the masked area, including
     intensity, shape, margin, or texture when these are observable.
  4. The position of the masked area relative to visible anatomy.
  5. Whether the masked area corresponds to a visually distinguishable
     finding. If it does not, state that explicitly.

The mask and metadata define the verified annotation target, but they
are not themselves visual evidence. Do not invent image features or
infer a diagnosis beyond the supplied metadata. If visual support is
limited, state this explicitly so the sample can be manually reviewed.
\end{verbatim}
\end{promptblock}

\paragraph{Prompt for normal samples.}\mbox{}\par
\begin{promptblock}
\begin{verbatim}
You are preparing an evidence-grounded description of a medical image.
You are given the image and verified annotation metadata.

Metadata:
  Modality: {modality}
  Anatomy: {anatomy}
  Status: normal

Use the metadata as annotation context, not as a substitute for visual
evidence. Inspect the image first, then write a concise structured
description covering:
  1. Imaging characteristics of the modality.
  2. The visible anatomical structures and their appearance.
  3. Whether the visible appearance is consistent with the source
     normality annotation.

Describe only what is visually supported. Do not introduce unobserved
findings. If a potentially abnormal feature is visible or the evidence
is ambiguous, state this explicitly so the sample can be manually
reviewed.
\end{verbatim}
\end{promptblock}

\subsubsection{Task-Specific CoT Generation}

For each VQA instance, Gemini 2.5 Pro receives the medical image, question,
answer options, source-derived correct answer, and auxiliary image description.
The correct answer is supplied as a fixed supervision target for rationale
synthesis, analogous to constructing an explanation for a verified annotation;
it is not used to measure model accuracy. To reduce post-hoc rationalization,
the instruction requires an observation-first derivation and requires any
insufficient or conflicting visual support to be stated explicitly for manual
review. We use a sampling temperature of 0.5 and a maximum generation length of
1,024 tokens, and two candidate reasoning traces are generated for each instance
before manual review.

\begin{promptblock}
\begin{verbatim}
You are preparing an evidence-grounded rationale for a verified medical
image annotation.

You are given a medical image, a multiple-choice question, the answer
options, the source-derived target answer, and an auxiliary image
description. The target answer is supervision for annotation synthesis;
it is not visual evidence. The image remains the primary evidence.

Image description: {description}
Question: {question}
Options: {options}
Target answer: {answer}

First derive the relevant observations from the image. Then explain how
those observations support the target answer.

Requirements:
  1. Begin with image observations before referring to the target.
  2. Address the specific decision required by the question.
  3. Refer only to visual evidence supported by the image.
  4. Do not treat the auxiliary description as authoritative when it
     conflicts with the image.
  5. Do not merely restate the target answer or invent evidence to make
     it appear correct.
  6. Prefer concise, task-relevant reasoning over generic medical text.
  7. If the visual evidence is insufficient or conflicts with the
     target, state this limitation explicitly rather than fabricating
     supporting evidence.

Task-specific focus: {task_focus}

Return the reasoning inside <think></think> tags, followed by the
target option inside <answer></answer> tags.
\end{verbatim}
\end{promptblock}

The task-specific focus is instantiated as follows: observable acquisition
characteristics consistent with the target modality for modality classification;
visible structures and anatomical landmarks supporting the target anatomy for
anatomy identification; visual evidence supporting the source normality label
for anomaly detection; the distribution of the finding supporting the derived
five-region label for lesion localization; and visible cellular, structural, or
morphological evidence compatible with the source-provided pathology label.

\subsubsection{CoT Annotation Statistics}

\begin{figure}[htbp]
    \centering
    \includegraphics[width=0.84\textwidth]{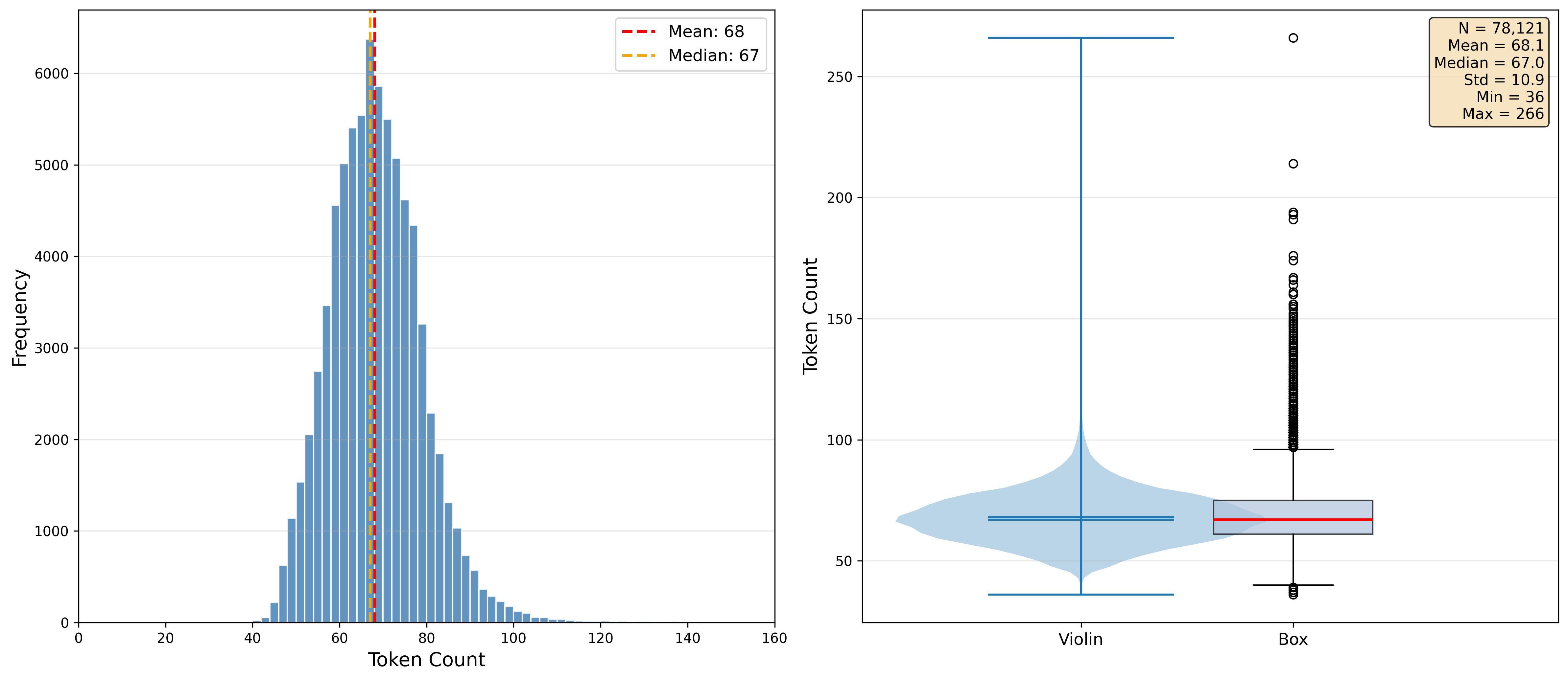}
    \caption{Overall token-count distribution of the generated CoT annotations.
    The histogram (left) marks the mean and median with dashed lines, while the
    violin and box plots (right) show the density, quartiles, and long-tail
    outliers. Token counts use the Qwen3.5 tokenizer and include only the text
    inside the \texttt{<think>...</think>} tags, excluding all tags and final
    answers. Across the complete training set ($N=78{,}121$), the mean is 68.1
    tokens, the median is 67.0, the standard deviation is 10.9, and the observed
    range is 36--266 tokens.}
    \label{fig:cot_token_distribution}
\end{figure}

\begin{figure}[htbp]
    \centering
    \includegraphics[width=0.84\textwidth]{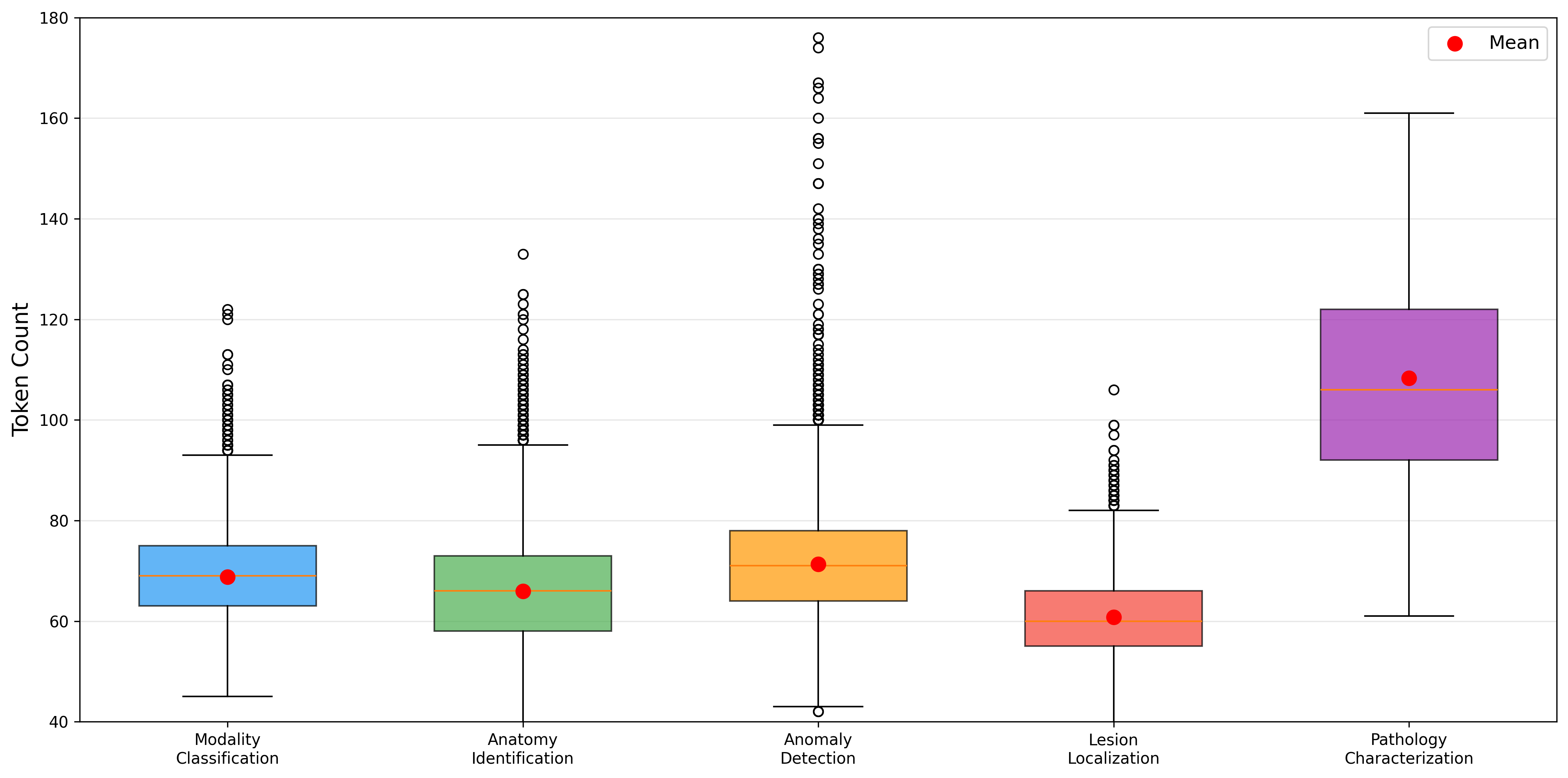}
    \caption{Token-count distributions of the CoT annotations for the five
    MedAD tasks. Boxes represent the interquartile ranges, horizontal lines
    indicate medians, and red markers indicate means. Pathology characterization
    requires the longest rationales on average, while lesion localization uses
    the shortest; anomaly detection exhibits the most pronounced upper-tail
    outliers.}
    \label{fig:cot_token_by_task}
\end{figure}

\begin{figure}[H]
    \centering
    \includegraphics[width=0.72\textwidth]{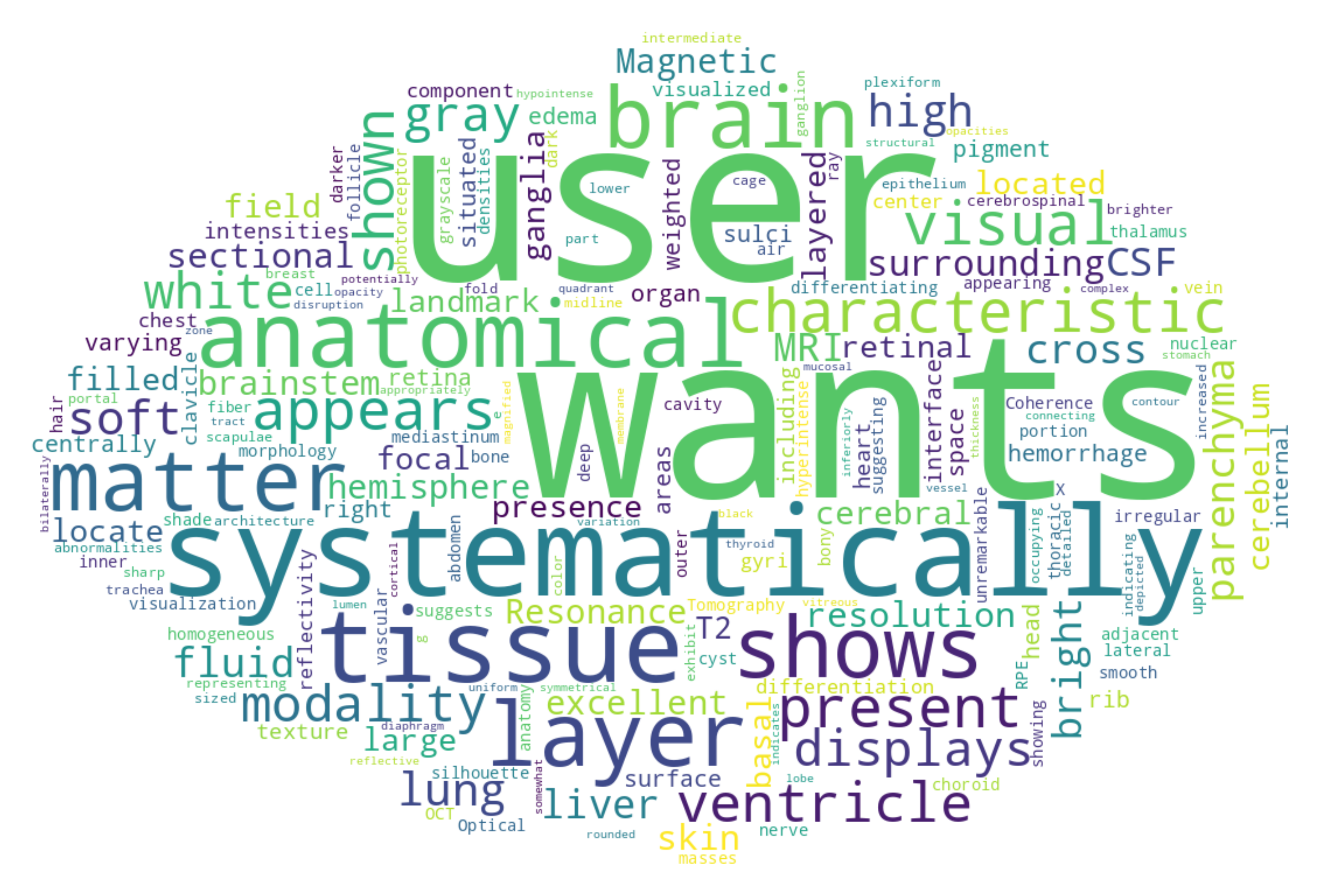}
    \caption{Word cloud of the generated CoT annotations, with word size
    proportional to corpus frequency. The dominant terms describe
    anatomy, imaging modalities, tissue appearance, visual evidence, and spatial
    relationships.}
    \label{fig:cot_wordcloud}
\end{figure}

Figures~\ref{fig:cot_token_distribution} and~\ref{fig:cot_token_by_task}
summarize the length characteristics of the generated CoT annotations, while
Figure~\ref{fig:cot_wordcloud} visualizes the annotation vocabulary. Token
counts are computed with the Qwen3.5 tokenizer using only the reasoning text
enclosed within the \texttt{<think>...</think>} tags. The literal
\texttt{<think>} and \texttt{</think>} delimiters, the
\texttt{<answer>...</answer>} tags, and the final answer text are excluded. The token counts are concentrated around 60--80 tokens: across all
78,121 training annotations, the mean and median are 68.1 and 67.0 tokens,
respectively. The task-level
distributions further show that pathology-characterization rationales are
substantially longer and more variable, whereas lesion-localization rationales
are generally the shortest. The word-frequency visualization indicates that the
annotations predominantly use image-centered terms describing anatomy,
modality, tissue appearance, and spatial or structural observations.

\clearpage

\subsection{Manual Quality Control}
\label{app:quality_control}

The review process is conducted by five master's students with medical
backgrounds. The workload is distributed across the annotators, and each
generated reasoning trace is assessed using the same five dimensions later
adopted by the Evidence-Aware Consistency Reward:
\begin{enumerate}\itemsep0pt
    \item \textbf{Visual grounding}: the reasoning is supported by visible image evidence;
    \item \textbf{Task relevance}: the reasoning directly addresses the question;
    \item \textbf{Answer support}: the reasoning logically supports the source-derived answer;
    \item \textbf{No hallucination}: the reasoning does not introduce unsupported medical findings;
    \item \textbf{No contradiction}: the reasoning is internally coherent and does not conflict with the final answer.
\end{enumerate}

The annotation pipeline operates as an iterative refinement loop over all
130,146 VQA prompts. The candidate traces generated for each prompt are
subjected to manual review, and one final trace is retained. Ground-truth answers remain fixed
throughout generation, review, revision, and regeneration and serve only as
supervision targets during annotation synthesis. Reviewers reject a trace even
when its final answer matches the target if the supporting evidence is absent,
hallucinated, or contradictory. Traces that fail review are revised when a
limited correction is sufficient or returned for regeneration; this process is
repeated until the trace satisfies all five quality criteria.

Through this multi-round process, 130,146 quality-controlled traces are retained
for the complete MedAD-38K dataset, including the 78,121 annotations associated
with the training split. As summarized in Table~\ref{tab:cot_disposition},
62.37\% of the final annotations are accepted without modification in the first
round, 8.45\% require limited manual revision, and 29.18\% are accepted after
iterative regeneration.

\begin{table}[htbp]
    \centering
    \setlength{\tabcolsep}{12pt}
    \renewcommand{\arraystretch}{1.1}
    \begin{tabular}{lrr}
        \toprule
        \textbf{Final Disposition Category} & \textbf{Traces} & \textbf{Proportion} \\
        \midrule
        Accepted without modification (Round 1) & 81,172 & 62.37\% \\
        Accepted after limited manual revision  & 10,997 & 8.45\% \\
        Accepted after iterative regeneration   & 37,977 & 29.18\% \\
        \midrule
        \textbf{Total Retained} & \textbf{130,146} & \textbf{100.00\%} \\
        \bottomrule
    \end{tabular}
    \caption{Composition of the final 130,146 retained CoT annotations for
    MedAD-38K after the iterative refinement process.}
    \label{tab:cot_disposition}
\end{table}

To quantify the reliability of the review criteria, we use stratified random
sampling to select 1,000 final CoT annotations covering all five MedAD tasks.
The subset contains 620 normal and 380 abnormal samples. Two annotators assess
each trace independently and blindly along the five dimensions above. Across
the resulting 5,000 dimension-level label pairs, their average agreement is
93.56\%. A third senior medical-imaging expert adjudicates every disagreement,
and the adjudicated labels are used as the final human reference.
Table~\ref{tab:cot_agreement} reports the reliability of the two independent
reviews before adjudication.

\begin{table}[htbp]
    \centering
    \setlength{\tabcolsep}{10pt}
    \renewcommand{\arraystretch}{1.1}
    \begin{tabular}{lccc}
        \toprule
        \textbf{Dimension} & \textbf{Agreement} & \textbf{Macro-F1} & \textbf{Cohen's $\kappa$} \\
        \midrule
        Visual Grounding & 91.20\% & 0.865 & 0.768 \\
        Task Relevance   & 95.60\% & 0.924 & 0.872 \\
        Answer Support   & 93.80\% & 0.895 & 0.821 \\
        No Hallucination & 92.50\% & 0.878 & 0.795 \\
        No Contradiction & 94.70\% & 0.912 & 0.854 \\
        \midrule
        \textbf{Average} & \textbf{93.56\%} & \textbf{0.895} & \textbf{0.822} \\
        \bottomrule
    \end{tabular}
    \caption{Inter-annotator agreement on 1,000 independently reviewed CoT
    annotations. Agreement is reported as a percentage; Macro-F1 and Cohen's
    $\kappa$ are unitless.}
    \label{tab:cot_agreement}
\end{table}

Agreement is at least 91.20\% for every dimension, and all Cohen's $\kappa$
values exceed 0.76, demonstrating reliable application of the five review
criteria. The third-expert adjudication then resolves all differences before
the annotations are retained as the final human reference.

\FloatBarrier

\subsection{Data Split and Leakage Prevention}
\label{app:data_split}

MedAD-38K is divided at the image level rather than at the VQA-instance level.
All questions derived from the same image remain in the same subset, so
different question templates or task types associated with one image cannot
appear across training and testing subsets.

A total of 40\% of the images are held out for testing. The remaining 60\% form
the training portion, which is further divided into disjoint SFT and RL subsets.
Approximately 70\% of the training images are used for SFT and 30\% are reserved
for RL. Table~\ref{tab:data_split} reports the resulting split.

\begin{table}[htbp]
    \centering
    \setlength{\tabcolsep}{10pt}
    \renewcommand{\arraystretch}{1.1}
    \begin{tabular}{lrrrr}
        \toprule
        \textbf{Split} & \textbf{Images} & \textbf{Image Ratio} &
        \textbf{VQA} & \textbf{VQA Ratio} \\
        \midrule
        Test      & 15,360 & 40.0\% & 52,025 & 40.0\% \\
        Train-SFT & 16,150 & 42.0\% & 54,709 & 42.0\% \\
        Train-RL  & 6,912  & 18.0\% & 23,412 & 18.0\% \\
        \midrule
        Training  & 23,062 & 60.0\% & 78,121 & 60.0\% \\
        Total     & 38,422 & 100.0\% & 130,146 & 100.0\% \\
        \bottomrule
    \end{tabular}
    \caption{Image-level split of MedAD-38K. All VQA instances associated with
    the same image remain in one subset.}
    \label{tab:data_split}
\end{table}

The SFT and RL subsets are mutually exclusive at the image level, and the
held-out test set is not used for supervised fine-tuning or policy optimization.
This protocol prevents the same medical image from appearing in different stages
through distinct questions or templates.

\FloatBarrier
\section{Reliability of the Language-Model Evaluators}
\label{app:evaluators}

Two distinct groups of language models are used in this work. MedGemma-4B
provides the reward signal during Con-GRPO, and a panel of three general-purpose
multimodal models performs the reasoning-answer consistency evaluation reported
in the main paper. This section documents the reliability of both groups and the
separation between them.

\subsection{Calibration of the Reward Evaluator}
\label{app:judge_calibration}

The Evidence-Aware Consistency Reward relies on an external evaluator
$\mathcal{F}$, instantiated as MedGemma-4B, to score the generated reasoning
along five binary dimensions. Since this evaluator provides the learning signal
for the reinforcement learning stage, its reliability directly affects the
resulting policy. An evaluator that systematically misinterprets medical
evidence would propagate its errors into training. We therefore calibrate the
evaluator against human annotations before using it for Con-GRPO.

\subsubsection{Annotation Protocol}

We construct a human-annotated reference set using the SFT-stage model. We
randomly sample 500 images from the training split while covering all five MedAD
tasks and all imaging modalities represented in MedAD-38K. For each image, we
draw six independent generations, yielding 3,000 reasoning-answer pairs.
Annotators with medical backgrounds label every pair along the same five binary
dimensions used by the reward.

\subsubsection{Development and Held-Out Evaluation}

The 3,000 pairs are divided into a development set of 2,000 pairs and a held-out
set of 1,000 pairs. Prompt refinement is performed only on the development set.
The held-out set is never inspected during this process and is used once, after
the prompt has been fixed.

With an initial instruction, the evaluator frequently disagrees with human
judgments, particularly on visual grounding and hallucination. We refine the
evaluation instruction on the development set so that it applies the five
criteria more consistently. This process is prompt calibration rather than model
fitting: it changes only the instruction given to the evaluator, and the
MedGemma-4B parameters remain frozen.

On the held-out set, the finalized prompt reaches an average agreement of
96.34\% with the human annotations, and every individual dimension exceeds
95\%. Because this measurement is taken on data that played no
role in prompt development, the agreement reflects the ability of the evaluator
to apply the intended criteria to unseen cases rather than to reproduce specific
annotated examples. During all reward computations, the sampling temperature is
fixed to zero so that the reward signal is deterministic and reproducible.
Table~\ref{tab:judge_calibration} reports the per-dimension results.

\begin{table}[htbp]
    \centering
    \setlength{\tabcolsep}{9pt}
    \renewcommand{\arraystretch}{1.1}
    \begin{tabular}{lcccc}
        \toprule
        \multirow{2}{*}{\textbf{Dimension}} &
        \multicolumn{2}{c}{\textbf{Initial Prompt}} &
        \multicolumn{2}{c}{\textbf{Calibrated Prompt}} \\
        \cmidrule(lr){2-3} \cmidrule(lr){4-5}
        & \textbf{Agreement} & \textbf{$\kappa$} & \textbf{Agreement} & \textbf{$\kappa$} \\
        \midrule
        Visual Grounding & 82.40\% & 0.615 & 95.80\% & 0.892 \\
        Task Relevance   & 86.50\% & 0.712 & 97.20\% & 0.935 \\
        Answer Support   & 84.10\% & 0.668 & 96.50\% & 0.914 \\
        No Hallucination & 81.20\% & 0.584 & 95.40\% & 0.885 \\
        No Contradiction & 85.30\% & 0.695 & 96.80\% & 0.921 \\
        \midrule
        \textbf{Average} & \textbf{83.90\%} & \textbf{0.655} &
        \textbf{96.34\%} & \textbf{0.909} \\
        \bottomrule
    \end{tabular}
    \caption{Agreement between MedGemma-4B and human annotations on the
    held-out calibration set, before and after prompt calibration.}
    \label{tab:judge_calibration}
\end{table}

Prompt calibration improves the average agreement by 12.44 percentage points
and the average Cohen's $\kappa$ by 0.254. The largest agreement gain occurs for
the no-hallucination criterion, which rises from 81.20\% to 95.40\%. These
results indicate that the calibrated evaluator applies all five criteria more
consistently while preserving a held-out separation between prompt development
and evaluation.

\FloatBarrier
\subsubsection{Reward Evaluator Prompt}

The evaluator receives the medical image, question, answer options, candidate
reasoning, and candidate answer. It does not receive the ground-truth label,
because answer correctness is scored separately by the accuracy reward.

\begin{promptblock}
\begin{verbatim}
You are a medical image reasoning evaluator.

You are given a medical image, a question, the answer options, a
candidate reasoning process, and the candidate final answer. Evaluate
the internal quality of the reasoning.

Do NOT judge whether the candidate answer is medically correct, and do
NOT compare it with any hidden ground-truth label. Assess only the
quality of the reasoning itself, given the image, the question, and the
answer the candidate produced.

Task type: {task_type}
Question: {question}
Options: {options}
Candidate reasoning: {reasoning}
Candidate final answer: {answer}

Score the candidate on five binary criteria.

1. visual_grounding
   1: the key visual evidence mentioned in the reasoning is supported
      by the medical image.
   0: the reasoning cites findings that are absent, unsupported, or so
      generic that they could apply to any image.

2. task_relevance
   1: the reasoning directly addresses the decision required by the
      question.
   0: the reasoning is off-topic, overly generic, or unrelated to the
      requested decision.

3. answer_support
   1: the reasoning logically supports the candidate answer.
   0: the reasoning does not support the answer, is too vague to
      support it, or supports a different option.

4. no_hallucination
   1: the reasoning contains no fabricated or unsupported medical
      finding.
   0: at least one obvious hallucinated finding is present.

5. no_contradiction
   1: the reasoning is internally coherent and does not conflict with
      the candidate answer.
   0: an internal contradiction or a conflict with the answer is
      present.

Return ONLY a JSON object, with no explanation and no markdown:
{
  "visual_grounding": 0 or 1,
  "task_relevance": 0 or 1,
  "answer_support": 0 or 1,
  "no_hallucination": 0 or 1,
  "no_contradiction": 0 or 1
}
\end{verbatim}
\end{promptblock}

\subsection{Separation Between Training and Evaluation Judges}
\label{app:judge_separation}

A policy trained against a particular evaluator may adapt to the preferences of
that evaluator. Measuring reasoning consistency with the same model that
provides the training reward would therefore introduce a risk of circular
evaluation, in which the reported improvement partly reflects the policy having
learned to satisfy its own judge.

To avoid this issue, the training and evaluation judges are kept entirely
separate. The reward evaluator used during Con-GRPO is MedGemma-4B, whereas the
consistency evaluation reported in the main paper is conducted by an independent
panel of three models from different families. MedGemma-4B takes no part in the
evaluation, and no evaluation judge provides a training reward.

The two roles also require different capabilities, which motivates the different
choices. The reward evaluator must inspect the medical image to determine
whether the candidate reasoning is visually grounded, so a domain-specific model
is appropriate. The consistency evaluation instead measures whether a reasoning
trace by itself entails a particular conclusion, which is primarily a question
of logical entailment rather than of specialized medical knowledge. Using
general-purpose models of independent origin for this role therefore provides a
stricter and less self-serving test.

\subsection{Consistency Evaluation Protocol}
\label{app:consistency_eval}

\subsubsection{Panel Composition and Metric Aggregation}

The evaluation panel consists of GLM-4.6V-Flash,
InternVL3.5-8B, and Qwen3-VL-8B-Instruct. Using models from different
families reduces reliance on the preferences of any single judge. The identity
of the evaluated model is hidden from the judges, all candidates are evaluated
with the same prompt and decoding configuration, and the sampling temperature
is fixed to zero.

All five inference passes of each fixed checkpoint are included in the
reasoning evaluation. Their outputs are pooled before metric aggregation,
yielding 260,125 MedAD-38K outputs and 91,260 external-set outputs per evaluated
checkpoint. Prompt A dimensions are averaged directly over all pooled outputs
and all three judges. For Prompt B, majority voting is first applied to each
pooled output, after which Consistency and Insuff. are computed as mean rates
over the resulting panel verdicts.

The panel produces two complementary evaluation outputs. First, to measure the
same five properties optimized by the Evidence-Aware Consistency Reward without
reusing its training-time evaluator, each judge independently applies the five
binary definitions in Equation~\ref{eq:eac_score} to the image, question,
options, reasoning, and model answer. For dimension $k$, the reported rate is
\begin{equation}
    M_k=\frac{100}{3N}\sum_{i=1}^{N}\sum_{j=1}^{3}d_{i,k}^{(j)},
\end{equation}
where $d_{i,k}^{(j)}\in\{0,1\}$ is judge $j$'s score. Thus, the five dimension
columns average the three judges' scores; no voting is used. These are
independent evaluation measurements rather than rewards produced by the
training-time MedGemma-4B evaluator.
The corresponding 0--5 aggregate is determined directly by
\begin{equation}
    \mathrm{Prompt\ A}=\frac{1}{100}\sum_{k=1}^{5}M_k.
\end{equation}

\subsubsection{Evaluation Prompt A: Five-Dimension Reasoning Quality}

Prompt A is image-conditioned and is applied independently by each of the
three evaluation judges. It uses the same five conceptual dimensions as the
Evidence-Aware Consistency Reward, but it is used only for evaluation and does
not provide a training signal.

\begin{promptblock}
\footnotesize
\begin{verbatim}
You are a medical image reasoning evaluator.

You are given a medical image, a question, the answer options, a
candidate reasoning process, and the candidate final answer. Evaluate
the internal quality of the reasoning.

Do NOT judge whether the candidate answer is medically correct, and do
NOT compare it with any hidden ground-truth label. Your task is to
assess the quality of the reasoning itself, given the image, the
question, and the answer the candidate produced.

Task type: {task_type}

Evaluation criteria (each scored 0 or 1):

1. visual_grounding
   - 1: The reasoning describes concrete visual evidence that is
        consistent with the medical image.
   - 0: The reasoning cites visual findings that are absent,
        unsupported, or only generic statements that could apply to
        any image.

2. task_relevance
   - 1: The reasoning directly addresses the diagnostic goal of the
        question.
   - 0: The reasoning does not address the diagnostic goal, or
        discusses aspects unrelated to what is being asked.

3. answer_support
   - 1: The reasoning clearly supports the candidate final answer.
   - 0: The reasoning does not support the candidate answer, is too
        vague to support it, or supports a different option.

4. no_hallucination
   - 1: The reasoning contains no fabricated or unsupported medical
        finding.
   - 0: The reasoning contains at least one obvious hallucinated or
        unsupported medical finding.

5. no_contradiction
   - 1: The reasoning is internally coherent and does not conflict
        with the candidate final answer.
   - 0: The reasoning contains an internal contradiction, or conflicts
        with the candidate final answer.

Question: {question}
Options: {options}
Candidate reasoning: {reasoning}
Candidate final answer: {answer}

Return ONLY a JSON object, with no explanation and no markdown
formatting:
{
  "visual_grounding": 0 or 1,
  "task_relevance": 0 or 1,
  "answer_support": 0 or 1,
  "no_hallucination": 0 or 1,
  "no_contradiction": 0 or 1
}
\end{verbatim}
\end{promptblock}

Second, reasoning--answer consistency is measured by the separate image-hidden
back-inference protocol below. Here each judge returns an inferred option, and
only these discrete inference results are combined by majority vote.

Each judge independently infers the answer supported by the candidate reasoning,
and the panel decision is determined by majority vote. If no answer option
receives at least two votes, the panel verdict is set to \texttt{INSUFFICIENT};
these samples remain in the denominator and are counted as inconsistent. To
quantify how well-determined these judgments are, we additionally report the pairwise
agreement and Fleiss' $\kappa$ among the three judges in
Table~\ref{tab:judge_agreement}.

The agreement analysis uses a separate pool of 1,000 generated outputs. We
randomly sample instances from the training data and obtain outputs from the
four SFT-stage backbone models: Qwen3.5-0.8B, InternVL3.5-1B, Qwen3-VL-2B, and
Qwen2.5-VL-3B. The same 1,000 outputs are then independently evaluated by all
three judges.

\begin{table}[htbp]
    \centering
    \setlength{\tabcolsep}{10pt}
    \renewcommand{\arraystretch}{1.1}
    \begin{tabular}{lrrr}
        \toprule
        \textbf{Judge Pair} & \textbf{Agreements} & \textbf{Samples} &
        \textbf{Rate} \\
        \midrule
        GLM-4.6V-Flash \textit{vs.} InternVL3.5-8B
            & 873 & 1,000 & 87.30\% \\
        GLM-4.6V-Flash \textit{vs.} Qwen3-VL-8B-Instruct
            & 909 & 1,000 & 90.90\% \\
        InternVL3.5-8B \textit{vs.} Qwen3-VL-8B-Instruct
            & 916 & 1,000 & 91.60\% \\
        \midrule
        \textbf{Average} & \textbf{2,698} & \textbf{3,000} &
            \textbf{89.93\%} \\
        \bottomrule
    \end{tabular}
    \caption{Agreement among the three evaluation judges on 1,000 outputs
    generated by the four SFT-stage backbone models from randomly sampled
    MedAD-38K training instances.}
    \label{tab:judge_agreement}
\end{table}

Across the 1,000 evaluated samples, all three judges agree on 857 samples
(85.70\%), exactly two judges agree on 127 samples (12.70\%), and all three
judges select different categories on 16 samples (1.60\%); the last group is
assigned the panel verdict \texttt{INSUFFICIENT}. The category counts
aggregated over the 3,000 individual judgments are 638 for A, 615 for B, 594
for C, 557 for D, and 596 for \texttt{INSUFFICIENT}. The observed mean
agreement is $\bar{P}=0.89933$, compared with a chance-expected agreement of
$P_e=0.20040$, yielding Fleiss' $\kappa=0.87410$ (87.41\%).

\subsubsection{Back-Inference Protocol}

Each judge receives the question, the answer options, and the generated
reasoning trace, but neither the medical image nor the final answer of the
evaluated model. Withholding the image is deliberate: it prevents a judge from
answering the question directly from visual evidence instead of inferring the
conclusion that the reasoning itself supports.

Let $\tilde{a}_{i}^{(j)}$ denote the answer inferred by judge $j$ for sample $i$.
The panel answer is
\begin{equation}
    \tilde{a}_{i} = \operatorname{MajorityVote}
    \left(\tilde{a}_{i}^{(1)},\, \tilde{a}_{i}^{(2)},\, \tilde{a}_{i}^{(3)}\right),
\end{equation}
and the reasoning-answer consistency indicator for sample $i$ is
\begin{equation}
    C_{i} = \mathbb{I}\left[\tilde{a}_{i} = a_{i}\right],
\end{equation}
where $a_{i}$ is the answer produced by the evaluated model. The consistency
rate over $N$ samples is
\begin{equation}
    \mathrm{Consistency} = \frac{1}{N}\sum_{i=1}^{N} C_{i}.
\end{equation}
The insufficient rate is computed from the final panel verdict,
\begin{equation}
    \mathrm{Insuff.}=\frac{100}{N}\sum_{i=1}^{N}
    \mathbb{I}[\tilde a_i=\texttt{INSUFFICIENT}],
\end{equation}
and therefore $100\,\mathrm{Consistency}+\mathrm{Insuff.}\leq100\%$.

Before evaluation, explicit restatements of the final option inside the
reasoning trace are removed by a rule-based filter. Without this step, a judge
could copy a stated conclusion instead of recovering it from the supporting
evidence. Judges may also return an \texttt{INSUFFICIENT} verdict when the
reasoning is too vague or self-contradictory to support any single option. Such
samples count toward $N$ but not toward the numerator, which avoids inflating
the consistency rate through forced guessing.

\subsubsection{Evaluation Prompt B: Back-Inference Consistency}

\begin{promptblock}
\begin{verbatim}
You are evaluating whether a reasoning process, on its own, leads to a
specific conclusion.

You are given a question, the answer options, and a reasoning process
written by a model. You are NOT given the model's final answer, and you
are NOT given the medical image.

Your task: based ONLY on the reasoning process, determine which option
the reasoning supports. Do not use outside medical knowledge to decide
which option is medically correct. Judge only what this particular
reasoning leads to.

If the reasoning explicitly states a final option, ignore that
statement and judge from the reasoning content alone.

If the reasoning is too vague, generic, or self-contradictory to
support any single option, output INSUFFICIENT.

Question: {question}
Options: {options}
Reasoning process: {reasoning}

Return ONLY a JSON object, with no explanation and no markdown
formatting:
{
  "inferred_option": "A" or "B" or "C" or "D" or "INSUFFICIENT"
}
\end{verbatim}
\end{promptblock}

\section{Extended Experimental Results}
\label{app:extended}

The main paper reports overall accuracy on MedAD-38K together with a single
aggregated consistency score. This section provides the detailed breakdowns that
do not fit within the page limit.

\begin{table}[htbp]
    \centering
    \setlength{\tabcolsep}{3pt}
    \begin{tabular}{lccccccc}
        \toprule
        \textbf{Model} & \textbf{Params} & \textbf{Anat.} & \textbf{Anom.} &
        \textbf{Loc.} & \textbf{Mod.} & \textbf{Path.} & \textbf{Overall} \\
        \midrule
        Qwen3.5-0.8B & 0.8B & 94.22$\pm$0.53 & 57.88$\pm$0.84 & 33.74$\pm$0.96 & 93.45$\pm$0.65 & 14.93$\pm$2.70 & 76.43$\pm$0.49 \\
        InternVL3.5-1B & 1B & 95.04$\pm$1.94 & 50.09$\pm$3.00 & 33.53$\pm$2.83 & 89.27$\pm$6.81 & 15.52$\pm$8.19 & 73.10$\pm$2.63 \\
        Qwen3VL-2B & 2B & 91.20$\pm$6.27 & 55.54$\pm$1.67 & 37.08$\pm$5.74 & 92.57$\pm$2.44 & 10.75$\pm$9.13 & 74.91$\pm$3.34 \\
        Qwen2.5VL-3B & 3B & 78.56$\pm$8.33 & 50.29$\pm$6.69 & 30.63$\pm$4.75 & 78.97$\pm$6.00 & 11.04$\pm$12.19 & 64.90$\pm$6.05 \\
        MiMo-VL-7B & 7B & 61.71$\pm$8.46 & 55.20$\pm$2.15 & 30.04$\pm$2.94 & 63.71$\pm$7.87 & 28.81$\pm$5.55 & 56.86$\pm$5.47 \\
        Qwen2.5VL-7B & 7B & 75.43$\pm$8.82 & 56.10$\pm$1.46 & 26.90$\pm$1.78 & 82.28$\pm$5.38 & 15.52$\pm$10.40 & 66.30$\pm$4.60 \\
        MiniCPM-V-4.5 & 9B & 90.01$\pm$3.27 & 54.65$\pm$3.39 & 26.64$\pm$1.17 & 93.24$\pm$2.02 & 6.04$\pm$4.22 & 73.37$\pm$2.62 \\
        Qwen3.5-9B & 9B & 94.39$\pm$0.22 & 61.27$\pm$1.34 & 34.40$\pm$2.30 & 96.11$\pm$0.16 & 20.90$\pm$2.99 & 78.38$\pm$0.36 \\
        GLM-4.6V-Flash & 9B & 87.55$\pm$3.44 & 58.04$\pm$4.26 & 48.21$\pm$13.50 & 87.11$\pm$6.33 & 16.12$\pm$8.62 & 74.15$\pm$4.71 \\
        Llama-3.2-Vision & 11B & 92.18$\pm$4.97 & 55.65$\pm$1.99 & 44.35$\pm$7.45 & 92.88$\pm$2.09 & 15.97$\pm$11.52 & 76.12$\pm$3.16 \\
        Gemma-4-26B & 26B & 93.80$\pm$1.04 & 65.78$\pm$1.11 & 36.58$\pm$1.68 & 94.28$\pm$2.97 & 12.76$\pm$5.85 & 79.18$\pm$1.43 \\
        Qwen3.6-27B & 27B & 93.82$\pm$0.26 & 60.16$\pm$0.61 & 39.71$\pm$0.59 & 90.67$\pm$0.35 & 8.36$\pm$1.50 & 76.76$\pm$0.32 \\
        Qwen3VL-32B & 32B & 84.62$\pm$11.15 & 53.48$\pm$8.60 & 37.83$\pm$9.18 & 90.44$\pm$6.12 & 18.43$\pm$8.47 & 71.84$\pm$8.28 \\
        InternVL3.5-38B & 38B & 94.13$\pm$0.93 & 62.06$\pm$1.53 & 50.52$\pm$4.38 & 94.67$\pm$1.08 & 17.80$\pm$7.16 & 79.79$\pm$1.32 \\
        \midrule
        MedVLM-R1$^{*}$ & 2B & 81.80$\pm$5.51 & 51.07$\pm$0.30 & 20.46$\pm$3.59 & 81.51$\pm$9.11 & 6.27$\pm$7.53 & 65.74$\pm$3.98 \\
        HuatuoGPT-Vision$^{*}$ & 7B & 84.09$\pm$9.33 & 45.46$\pm$12.65 & 34.08$\pm$4.74 & 85.52$\pm$8.28 & 4.70$\pm$3.24 & 67.38$\pm$7.96 \\
        Lingshu-7B$^{*}$ & 7B & 80.44$\pm$6.26 & 53.03$\pm$4.51 & 29.31$\pm$6.07 & 76.52$\pm$5.40 & 12.84$\pm$3.90 & 65.41$\pm$4.45 \\
        LLaVA-Med$^{*}$ & 7B & 76.85$\pm$6.96 & 50.10$\pm$0.99 & 22.35$\pm$5.28 & 74.90$\pm$10.85 & 7.16$\pm$4.61 & 62.24$\pm$4.44 \\
        \midrule
        InternVL3.5-8B$^{\dagger}$ & 8B & 95.77$\pm$1.15 & 76.66$\pm$3.17 & 83.18$\pm$4.89 & \underline{96.16}$\pm$0.38 & 27.27$\pm$5.06 & 88.53$\pm$1.04 \\
        Qwen3VL-8B$^{\dagger}$ & 8B & 95.87$\pm$1.65 & 79.25$\pm$8.12 & 70.11$\pm$12.67 & 95.53$\pm$1.73 & 72.46$\pm$7.89 & 88.01$\pm$4.02 \\
        Qwen3.6-27B$^{\dagger}$ & 27B & \underline{95.91}$\pm$1.42 & \underline{87.38}$\pm$1.41 & \underline{91.44}$\pm$0.73 & 95.61$\pm$1.60 & \underline{82.11}$\pm$1.02 & \underline{92.75}$\pm$0.52 \\
        \midrule
        \textbf{MedAD-R1 (Ours)} & 0.8B & \textbf{98.96}$\pm$0.70 & \textbf{88.30}$\pm$0.10 & \textbf{91.49}$\pm$1.20 & \textbf{99.57}$\pm$0.08 & \textbf{85.57}$\pm$1.07 & \textbf{95.12}$\pm$0.04 \\
        \bottomrule
    \end{tabular}
    \caption{Detailed MedAD-38K test accuracy over five independent inference passes of each fixed checkpoint, reported as mean$\pm$standard deviation (\%). No retraining is performed between passes. Overall is micro-averaged over all applicable VQA instances. $^{*}$ marks medical LMMs and $^{\dagger}$ marks larger models fine-tuned on our training data. Boldface denotes the best result and underlining the second best.}
    \label{tab:indomain_accuracy_detailed}
\end{table}

\subsection{Detailed Reasoning-Quality Results}
\label{app:per_dimension}

Table~\ref{tab:consistency_test_results} reports the five reasoning-quality
dimensions produced by Prompt A, its aggregate score on a 0--5 scale,
reasoning--answer consistency from Prompt B, and the proportion of outputs
assigned the final panel verdict \texttt{INSUFFICIENT} on the held-out
MedAD-38K test set. The five dimension columns average the three independent
judge scores according to Section~\ref{app:consistency_eval}; only Consist. and
Insuff. derive from the majority-voted Prompt-B results.

\begin{table}[htbp]
    \centering
    \setlength{\tabcolsep}{1.4pt}
    \begin{tabular}{lccccccccc}
        \toprule
        \textbf{Model} & \textbf{Params} & \textbf{Vis. Ground.} & \textbf{Task Rel.} & \textbf{Ans. Supp.} & \textbf{No Hall.} & \textbf{No Contr.} & \textbf{Prompt A} & \textbf{Consist.} & \textbf{Insuff.} \\
        \midrule
        Qwen3.5-0.8B & 0.8B & 78.56 & 98.82 & 81.45 & 96.83 & 97.54 & 4.53 & 92.12 & 4.45 \\
        InternVL3.5-1B & 1B & 76.24 & 97.63 & 77.82 & 96.87 & 96.15 & 4.45 & 93.53 & 0.45 \\
        Qwen3-VL-2B & 2B & 80.56 & 97.68 & 82.15 & 96.82 & 97.53 & 4.55 & 94.84 & 0.42 \\
        Qwen2.5-VL-3B & 3B & 60.15 & 96.54 & 62.52 & 96.91 & 97.23 & 4.13 & 93.25 & 2.74 \\
        MiMo-VL-7B & 7B & 50.24 & 96.51 & 52.49 & 94.56 & 95.55 & 3.89 & 94.57 & 0.47 \\
        Qwen2.5-VL-7B & 7B & 75.42 & 97.83 & 76.68 & 97.16 & 97.66 & 4.45 & 94.17 & 3.22 \\
        MiniCPM-V-4.5 & 9B & 86.45 & 98.84 & 87.18 & 97.35 & 98.16 & 4.68 & 94.96 & 0.51 \\
        Qwen3.5-9B & 9B & 88.75 & 99.75 & 89.42 & 96.90 & 98.73 & 4.74 & 94.25 & 0.53 \\
        GLM-4.6V-Flash & 9B & 72.18 & 97.94 & 73.24 & 97.21 & 97.15 & 4.38 & 93.43 & 0.46 \\
        Llama-3.2-Vision & 11B & 84.23 & 98.92 & 85.73 & 97.91 & 98.62 & 4.65 & 94.04 & 0.51 \\
        Gemma-4-26B & 26B & 94.92 & 99.78 & 95.17 & 98.52 & 99.34 & 4.88 & 96.95 & 0.18 \\
        Qwen3.6-27B & 27B & 89.24 & 99.72 & 89.54 & 97.10 & 98.45 & 4.74 & 92.22 & 0.37 \\
        Qwen3-VL-32B & 32B & 93.85 & 99.87 & 94.23 & 98.41 & 98.02 & 4.84 & 96.23 & 0.12 \\
        InternVL3.5-38B & 38B & 83.45 & 99.68 & 84.05 & 97.16 & 98.74 & 4.63 & 94.00 & 0.23 \\
        \midrule
        MedVLM-R1$^{*}$ & 2B & 51.45 & 96.40 & 52.40 & 97.10 & 97.50 & 3.95 & 38.90 & 54.90 \\
        HuatuoGPT-Vision$^{*}$ & 7B & 65.32 & 96.70 & 66.80 & 97.40 & 97.30 & 4.24 & 32.50 & 58.90 \\
        Lingshu-7B$^{*}$ & 7B & 77.35 & 98.72 & 78.70 & 97.30 & 98.70 & 4.51 & 95.60 & 0.14 \\
        LLaVA-Med$^{*}$ & 7B & 52.18 & 96.60 & 53.00 & 97.40 & 97.30 & 3.96 & 31.50 & 66.31 \\
        InternVL3.5-8B$^{\dagger}$ & 8B & 94.65 & 99.86 & 95.13 & 98.74 & 99.65 & 4.88 & 95.39 & 0.15 \\
        Qwen3VL-8B$^{\dagger}$ & 8B & 94.12 & 99.85 & 93.85 & 98.54 & 99.48 & 4.86 & 93.33 & 0.19 \\
        Qwen3.6-27B$^{\dagger}$ & 27B & 91.15 & 99.91 & 91.77 & 97.83 & 99.25 & 4.80 & 92.10 & 0.18 \\
        \midrule
        SFT only (Qwen3.5-0.8B) & 0.8B & 95.15 & 99.92 & 95.43 & 98.22 & 99.47 & 4.88 & 94.35 & 0.24 \\
        SFT only (InternVL3.5-1B) & 1B & 94.68 & 99.88 & 94.85 & 98.15 & 99.31 & 4.87 & 93.75 & 0.25 \\
        SFT only (Qwen3VL-2B) & 2B & 94.25 & 99.87 & 94.51 & 97.98 & 99.12 & 4.86 & 92.84 & 0.21 \\
        SFT only (Qwen2.5VL-3B) & 3B & 93.82 & 99.85 & 94.08 & 97.70 & 99.06 & 4.85 & 92.84 & 0.28 \\
        \midrule
        MedAD-R1 (Ours) & 0.8B & \textbf{97.42} & \textbf{99.97} & \textbf{97.53} & \textbf{99.34} & \textbf{99.68} & \textbf{4.94} & \textbf{97.26} & 0.08 \\
        MedAD-R1 (InternVL3.5-1B) & 1B & 96.85 & 99.95 & 97.12 & 99.24 & 99.53 & 4.93 & 96.73 & 0.06 \\
        MedAD-R1 (Qwen3VL-2B) & 2B & 96.34 & 99.96 & 96.46 & 99.15 & 99.41 & 4.91 & 96.84 & \textbf{0.05} \\
        MedAD-R1 (Qwen2.5VL-3B) & 3B & 95.78 & 99.94 & 95.92 & 98.90 & 99.36 & 4.90 & 96.15 & \textbf{0.05} \\
        \bottomrule
    \end{tabular}
    \caption{Main results on the pooled outputs from all five MedAD-38K inference passes. The five dimensions average the three independent judge scores, and Prompt A is their sum on a 0--5 scale; Consist. and Insuff. use the majority-voted Prompt-B verdict. SFT-only rows are the Stage-1 counterparts of the four complete MedAD-R1 models. Vis. Ground.: Visual Grounding; Task Rel.: Task Relevance; Ans. Supp.: Answer Support; No Hall.: No Hallucination; No Contr.: No Contradiction. All metrics except Prompt A are percentages. Higher is better except for Insuff. Boldface marks the best value in each column. $^{*}$ marks a medical LMM and $^{\dagger}$ a larger model fine-tuned on our data.}
    \label{tab:consistency_test_results}
\end{table}

\FloatBarrier
\subsection{External Datasets}
\label{app:external_data}

Beyond the held-out MedAD-38K test set, we evaluate zero-shot generalization on
five source-disjoint external datasets: ColonDB, HeadCT, EndoTect (Endo), TN3K,
and the Non-Melanoma Skin Cancer Histopathology (NMSC) dataset. None of these datasets
contributes images to MedAD-38K or to either training stage.

For each external dataset, we construct VQA instances under the same unified
MedAD protocol described in Section~\ref{app:vqa_construction}. Questions are
generated only for the tasks supported by the original annotations of each
dataset, and lesion-localization labels are derived from source-provided masks
using the five-region procedure of Section~\ref{app:five_region}. No CoT
annotations are generated, because these datasets are used only for zero-shot
evaluation.

\begin{table}[htbp]
    \centering
    \setlength{\tabcolsep}{9pt}
    \renewcommand{\arraystretch}{1.1}
    \begin{tabular}{lllrr}
        \toprule
        \textbf{Dataset} & \textbf{Modality} & \textbf{Anatomy} &
        \textbf{Images} & \textbf{VQA} \\
        \midrule
        ColonDB             & Endoscopy  & Gastrointestinal & 380   & 1,520 \\
        HeadCT              & CT         & Brain            & 200   & 800 \\
        EndoTect (Endo)     & Endoscopy  & Gastrointestinal & 200   & 800 \\
        TN3K                & Ultrasound & Thyroid          & 3,493 & 13,972 \\
        NMSC Histopathology & Microscopy & Skin             & 290   & 1,160 \\
        \midrule
        \textbf{Total} & & & \textbf{4,563} & \textbf{18,252} \\
        \bottomrule
    \end{tabular}
    \caption{Composition of the five source-disjoint external evaluation
    datasets.}
    \label{tab:external_stats}
\end{table}

Across the external collection, anatomy identification, anomaly detection, and
modality classification each contain 4,563 instances; lesion localization
contains 4,456 instances; and pathology characterization contains 107 instances.
These counts sum to 18,252 and are the weights used for the micro-averaged
Overall accuracy.

Table~\ref{tab:external_results} reports the complete task-level accuracy
results on the external test set. Table~\ref{tab:consistency_etenl_results}
reports the corresponding reasoning-quality and consistency results. The
compact table in the main paper retains only EOverall and EConsist.

\begin{table}[htbp]
    \centering
    \small
    \setlength{\tabcolsep}{2.8pt}
    \begin{tabular}{lccccccc}
        \toprule
        \textbf{Model} & \textbf{Params} & \textbf{Anatomy ID.} & \textbf{Anomaly Det.} & \textbf{Lesion Loc.} & \textbf{Modality Class.} & \textbf{Pathology Char.} & \textbf{Overall} \\
        \midrule
        Qwen3.5-0.8B & 0.8B & 72.80$\pm$0.97 & 55.80$\pm$4.87 & 43.94$\pm$0.98 & 99.29$\pm$0.27 & 71.20$\pm$6.79 & 68.12$\pm$1.25 \\
        InternVL3.5-1B & 1B & 89.26$\pm$2.94 & 55.82$\pm$7.12 & 37.40$\pm$5.57 & 95.73$\pm$7.52 & 81.20$\pm$6.11 & 69.81$\pm$4.95 \\
        Qwen3-VL-2B & 2B & 85.88$\pm$10.10 & 42.60$\pm$3.01 & 42.26$\pm$9.59 & 98.21$\pm$1.77 & 62.00$\pm$17.03 & 67.35$\pm$5.53 \\
        Qwen2.5-VL-3B & 3B & 45.16$\pm$5.73 & 39.45$\pm$7.26 & 31.40$\pm$5.51 & 87.60$\pm$3.75 & 60.40$\pm$15.81 & 51.07$\pm$4.79 \\
        MiMo-VL-7B & 7B & 42.79$\pm$11.25 & 39.61$\pm$7.48 & 26.31$\pm$1.23 & 48.50$\pm$7.87 & 58.00$\pm$19.06 & 39.49$\pm$5.86 \\
        Qwen2.5-VL-7B & 7B & 51.31$\pm$10.54 & 34.87$\pm$8.69 & 27.22$\pm$2.18 & 85.77$\pm$7.91 & 82.20$\pm$11.44 & 50.12$\pm$5.88 \\
        MiniCPM-V-4.5 & 9B & 61.09$\pm$6.11 & 51.70$\pm$6.44 & 27.38$\pm$2.01 & 95.77$\pm$2.54 & 86.60$\pm$13.48 & 59.33$\pm$4.14 \\
        Qwen3.5-9B & 9B & 80.16$\pm$1.47 & 69.49$\pm$1.94 & 53.41$\pm$3.88 & 99.85$\pm$0.11 & 95.20$\pm$2.23 & 75.97$\pm$1.12 \\
        GLM-4.6V-Flash & 9B & 71.66$\pm$9.85 & 52.32$\pm$4.19 & 49.23$\pm$17.16 & 92.98$\pm$5.87 & 91.80$\pm$6.85 & 66.80$\pm$8.87 \\
        Llama-3.2-Vision & 11B & 83.86$\pm$10.09 & 42.12$\pm$2.97 & 49.95$\pm$11.87 & 98.51$\pm$1.37 & 64.40$\pm$18.62 & 68.70$\pm$5.90 \\
        Gemma-4-26B & 26B & 87.77$\pm$4.71 & 69.86$\pm$2.15 & 48.21$\pm$6.86 & 99.12$\pm$0.79 & 92.40$\pm$7.17 & 76.50$\pm$3.26 \\
        Qwen3.6-27B & 27B & 78.30$\pm$1.03 & 70.14$\pm$0.46 & 60.07$\pm$0.89 & \textbf{99.89$\pm$0.04} & 87.60$\pm$2.87 & 77.26$\pm$0.41 \\
        Qwen3-VL-32B & 32B & 66.89$\pm$20.27 & 64.87$\pm$13.29 & 49.09$\pm$19.18 & 93.34$\pm$6.59 & 79.00$\pm$17.20 & 68.73$\pm$14.78 \\
        InternVL3.5-38B & 38B & 92.43$\pm$5.97 & 61.44$\pm$3.14 & 64.94$\pm$11.15 & 99.32$\pm$0.49 & \textbf{97.40$\pm$3.77} & 79.72$\pm$4.49 \\
        \midrule
        MedVLM-R1$^{*}$ & 2B & 46.22$\pm$3.72 & 57.44$\pm$10.41 & 21.73$\pm$1.85 & 91.64$\pm$7.95 & 75.20$\pm$22.64 & 54.57$\pm$4.01 \\
        HuatuoGPT-Vision$^{*}$ & 7B & 78.28$\pm$14.15 & 50.46$\pm$16.25 & 31.42$\pm$4.28 & 89.96$\pm$9.65 & 95.40$\pm$3.50 & 62.90$\pm$9.83 \\
        Lingshu-7B$^{*}$ & 7B & 59.31$\pm$21.77 & 43.47$\pm$10.46 & 29.49$\pm$9.50 & 83.21$\pm$18.46 & 70.20$\pm$29.11 & 54.11$\pm$13.74 \\
        LLaVA-Med$^{*}$ & 7B & 46.60$\pm$4.71 & 62.39$\pm$15.16 & 22.82$\pm$3.30 & 87.93$\pm$9.85 & 76.20$\pm$18.12 & 55.25$\pm$5.70 \\
        InternVL3.5-8B$^{\dagger}$ & 8B & 81.25$\pm$1.16 & 86.74$\pm$7.09 & 67.69$\pm$3.45 & 98.09$\pm$0.30 & 86.20$\pm$3.82 & 83.55$\pm$2.46 \\
        Qwen3VL-8B$^{\dagger}$ & 8B & 67.99$\pm$3.32 & 86.36$\pm$15.76 & 52.04$\pm$17.60 & 96.07$\pm$0.95 & 28.60$\pm$4.76 & 75.48$\pm$9.12 \\
        Qwen3.6-27B$^{\dagger}$ & 27B & 66.59$\pm$1.76 & \textbf{99.68$\pm$0.03} & 73.92$\pm$0.11 & 97.31$\pm$0.04 & 55.40$\pm$2.24 & 84.27$\pm$0.40 \\
        \midrule
        MedAD-R1 (Ours) & 0.8B & 86.53$\pm$1.75 & 94.57$\pm$1.90 & \textbf{82.42$\pm$1.34} & 97.44$\pm$0.57 & 57.33$\pm$7.41 & \textbf{90.09$\pm$1.12} \\
        MedAD-R1 (InternVL3.5-1B) & 1B & 88.67$\pm$2.57 & 79.52$\pm$3.06 & 70.21$\pm$1.12 & 96.67$\pm$0.87 & 61.33$\pm$15.33 & 83.72$\pm$0.66 \\
        MedAD-R1 (Qwen3VL-2B) & 2B & \textbf{95.28$\pm$0.82} & 97.26$\pm$1.09 & 63.14$\pm$10.82 & 95.55$\pm$1.53 & 36.67$\pm$15.43 & 87.65$\pm$2.68 \\
        MedAD-R1 (Qwen2.5VL-3B) & 3B & 95.26$\pm$0.38 & 81.13$\pm$3.97 & 70.24$\pm$1.79 & 96.18$\pm$0.17 & 26.33$\pm$1.89 & 85.44$\pm$0.84 \\
        \bottomrule
    \end{tabular}
    \caption{External-test accuracy over five independent inference passes of each fixed checkpoint, reported as mean$\pm$standard deviation (\%). No retraining is performed between passes. Overall is micro-averaged across all 18,252 applicable VQA instances rather than computed as an unweighted mean of the task columns. Boldface marks the best value in each column. $^{*}$ marks a medical LMM and $^{\dagger}$ a larger model fine-tuned on our data.}
    \label{tab:external_results}
\end{table}

\begin{table}[htbp]
    \centering
    \setlength{\tabcolsep}{1.4pt}
    \begin{tabular}{lccccccccc}
        \toprule
        \textbf{Model} & \textbf{Params} & \textbf{Vis. Ground.} & \textbf{Task Rel.} & \textbf{Ans. Supp.} & \textbf{No Hall.} & \textbf{No Contr.} & \textbf{Prompt A} & \textbf{Consist.} & \textbf{Insuff.} \\
        \midrule
        Qwen3.5-0.8B & 0.8B & 72.34 & 97.72 & 74.19 & 97.41 & 98.20 & 4.40 & 88.75 & 6.76 \\
        InternVL3.5-1B & 1B & 68.96 & 98.96 & 70.17 & 97.09 & 96.14 & 4.31 & 91.53 & 0.22 \\
        Qwen3-VL-2B & 2B & 78.24 & 98.93 & 79.25 & 97.29 & 98.65 & 4.52 & 94.41 & 0.18 \\
        Qwen2.5-VL-3B & 3B & 55.17 & 97.55 & 56.89 & 96.91 & 98.33 & 4.05 & 92.01 & 2.33 \\
        MiMo-VL-7B & 7B & 46.93 & 97.50 & 48.58 & 95.84 & 97.92 & 3.87 & 93.78 & 0.13 \\
        Qwen2.5-VL-7B & 7B & 69.30 & 98.79 & 70.39 & 98.07 & 98.72 & 4.35 & 91.84 & 3.79 \\
        MiniCPM-V-4.5 & 9B & 85.42 & 98.90 & 86.23 & 98.35 & 98.46 & 4.67 & 95.10 & 0.43 \\
        Qwen3.5-9B & 9B & 86.35 & 99.80 & 87.51 & 96.76 & 99.16 & 4.70 & 95.05 & 0.35 \\
        GLM-4.6V-Flash & 9B & 66.29 & 98.97 & 67.90 & 97.86 & 98.53 & 4.30 & 94.43 & \textbf{0.05} \\
        Llama-3.2-Vision & 11B & 84.75 & 98.89 & 85.67 & 97.95 & 98.61 & 4.66 & 95.67 & 0.15 \\
        Gemma-4-26B & 26B & 90.85 & 99.89 & 91.49 & 98.60 & 99.24 & 4.80 & 96.01 & 0.13 \\
        Qwen3.6-27B & 27B & 86.42 & 99.81 & 87.27 & 97.23 & 98.52 & 4.69 & 88.93 & 0.25 \\
        Qwen3-VL-32B & 32B & 89.85 & 99.87 & 90.12 & 98.38 & 98.77 & 4.77 & 95.66 & \textbf{0.05} \\
        InternVL3.5-38B & 38B & 80.89 & 99.79 & 81.53 & 96.44 & 98.81 & 4.57 & 94.49 & 0.08 \\
        \midrule
        MedVLM-R1$^{*}$ & 2B & 48.32 & 96.16 & 49.41 & 97.87 & 97.30 & 3.89 & 36.85 & 56.67 \\
        HuatuoGPT-Vision$^{*}$ & 7B & 60.39 & 97.66 & 61.84 & 98.28 & 98.12 & 4.16 & 29.25 & 60.64 \\
        Lingshu-7B$^{*}$ & 7B & 74.55 & 99.77 & 75.12 & 98.29 & 98.70 & 4.46 & 94.03 & 0.06 \\
        LLaVA-Med$^{*}$ & 7B & 49.75 & 97.46 & 50.62 & 98.15 & 97.40 & 3.93 & 28.40 & 69.31 \\
        InternVL3.5-8B$^{\dagger}$ & 8B & 92.35 & 99.90 & 93.13 & 98.55 & \textbf{99.67} & 4.84 & 94.67 & 0.18 \\
        Qwen3VL-8B$^{\dagger}$ & 8B & 90.85 & 99.85 & 91.45 & 97.54 & 99.41 & 4.79 & 91.92 & 0.23 \\
        Qwen3.6-27B$^{\dagger}$ & 27B & 88.15 & 99.87 & 89.20 & 97.53 & 99.50 & 4.74 & 90.94 & 0.27 \\
        \midrule
        SFT only (Qwen3.5-0.8B) & 0.8B & 91.45 & 99.88 & 91.83 & 97.60 & 99.25 & 4.80 & 92.48 & 0.24 \\
        SFT only (InternVL3.5-1B) & 1B & 90.82 & 99.86 & 91.21 & 96.55 & 99.16 & 4.78 & 91.41 & 0.28 \\
        SFT only (Qwen3VL-2B) & 2B & 90.25 & 99.85 & 90.69 & 96.61 & 98.99 & 4.76 & 90.54 & 0.21 \\
        SFT only (Qwen2.5VL-3B) & 3B & 89.65 & 99.82 & 90.11 & 95.22 & 99.10 & 4.74 & 89.60 & 0.32 \\
        \midrule
        MedAD-R1 (Ours) & 0.8B & \textbf{94.25} & \textbf{99.95} & \textbf{94.60} & \textbf{98.88} & 99.47 & \textbf{4.87} & \textbf{96.26} & 0.12 \\
        MedAD-R1 (InternVL3.5-1B) & 1B & 93.42 & 99.92 & 93.80 & 98.72 & 99.30 & 4.85 & 94.33 & 0.15 \\
        MedAD-R1 (Qwen3VL-2B) & 2B & 92.85 & 99.94 & 93.15 & 98.50 & 99.25 & 4.84 & 94.56 & 0.10 \\
        MedAD-R1 (Qwen2.5VL-3B) & 3B & 92.15 & 99.91 & 92.54 & 98.30 & 99.25 & 4.82 & 94.15 & 0.14 \\
        \bottomrule
    \end{tabular}
    \caption{Reasoning-quality results on the pooled outputs from all five external-set inference passes. The five dimensions average the three independent judge scores, and Prompt A is their sum on a 0--5 scale; Consist. and Insuff. use the majority-voted Prompt-B verdict. SFT-only rows are the Stage-1 counterparts of the four complete MedAD-R1 models. The abbreviations and symbols follow Table~\ref{tab:consistency_test_results}; all metrics except Prompt A are percentages, and lower is better only for Insuff.}
    \label{tab:consistency_etenl_results}
\end{table}

\FloatBarrier
\subsection{Cross-Backbone Results}
\label{app:cross_backbone}

Qwen3.5-0.8B is the primary MedAD-R1 backbone. The same two-stage framework is
also applied to InternVL3.5-1B, Qwen3-VL-2B, and Qwen2.5-VL-3B to test whether
the improvement transfers across model families and scales.
Table~\ref{tab:cross_backbone} reports the corresponding zero-shot, SFT-only,
and Con-GRPO results. The rightmost column gives the improvement of Con-GRPO
over the SFT-only model with the same backbone.

\begin{table}[htbp]
    \centering
    \setlength{\tabcolsep}{6.8pt}
    \renewcommand{\arraystretch}{1.1}
    \begin{tabular}{llccccccc}
        \toprule
        \textbf{Backbone} & \textbf{Setting} & \textbf{Anatomy} &
        \textbf{Modality} & \textbf{Anomaly} & \textbf{Localization} &
        \textbf{Pathology} & \textbf{Overall} & \textbf{$\Delta$} \\
        \midrule
        \multirow{3}{*}{Qwen3.5-0.8B}
            & Zero-shot & 94.22 & 93.45 & 57.88 & 33.74 & 14.93 & 76.43 & -- \\
            & SFT only  & 95.11 & 95.07 & 82.73 & 90.68 & 81.09 & 90.89 & -- \\
            & Con-GRPO  & 98.96 & \textbf{99.57} & \textbf{88.30} & \textbf{91.49} & 85.57 & \textbf{95.12} & +4.23 \\
        \midrule
        \multirow{3}{*}{InternVL3.5-1B}
            & Zero-shot & 95.04 & 89.27 & 50.09 & 33.53 & 15.52 & 73.10 & -- \\
            & SFT only  & 92.58 & 91.51 & 76.24 & 83.61 & 79.10 & 86.40 & -- \\
            & Con-GRPO  & 97.88 & 98.51 & 81.40 & 90.45 & 80.85 & 92.31 & \textbf{+5.91} \\
        \midrule
        \multirow{3}{*}{Qwen3-VL-2B}
            & Zero-shot & 91.20 & 92.57 & 55.54 & 37.08 & 10.75 & 74.91 & -- \\
            & SFT only  & 95.98 & 95.03 & 74.11 & 75.72 & 26.37 & 86.72 & -- \\
            & Con-GRPO  & 97.01 & 91.66 & 86.83 & 85.40 & 29.36 & 90.83 & +4.11 \\
        \midrule
        \multirow{3}{*}{Qwen2.5-VL-3B}
            & Zero-shot & 78.56 & 78.97 & 50.29 & 30.63 & 11.04 & 64.90 & -- \\
            & SFT only  & 95.94 & 96.22 & 72.06 & 88.22 & 65.67 & 87.97 & -- \\
            & Con-GRPO  & \textbf{99.01} & 99.49 & 81.11 & 90.16 & \textbf{86.07} & 92.85 & +4.88 \\
        \bottomrule
    \end{tabular}
    \caption{Per-task accuracy for the four backbones under zero-shot, SFT-only,
    and complete Con-GRPO training. Overall is micro-averaged over all applicable
    test instances rather than computed as an unweighted mean of the task columns.
    Boldface marks the best value in each metric column.}
    \label{tab:cross_backbone}
\end{table}

\subsection{Implementation Details}
\label{app:implementation}

\subsubsection{Hardware and Software Environment}

The main training experiments were conducted on a server equipped with eight
NVIDIA A100-SXM4-80GB GPUs, an Intel Xeon Platinum 8369B CPU, and 800~GiB of
system memory. Six GPUs were used for Distributed Data Parallel (DDP) policy
training, while the remaining two GPUs served the MedGemma-4B reward evaluator
during Con-GRPO. All training used BFloat16 mixed precision.

The training environment used CUDA 12.8, Python 3.10.13, PyTorch 2.10.0,
Hugging Face Transformers 5.3.0, ms-swift 4.5.0.dev0, PEFT 0.19.1,
TRL 0.29.1, and Accelerate 1.14.0. This environment is the primary experimental
environment used to train and evaluate the models. It is distinct from the
controlled efficiency-measurement environment described in
Section~\ref{app:efficiency}.

\subsubsection{Model Architecture and LoRA Configuration}

The primary backbone is reported using its official Qwen3.5-0.8B model
designation. We apply Low-Rank Adaptation (LoRA) to all
linear layers in the language model, with rank $r=16$, scaling parameter
$\alpha=32$, and dropout 0.05. Only the LoRA adapter parameters are optimized;
the underlying backbone parameters remain frozen.

The vision encoder and vision--language aligner remain frozen throughout both
training stages so that adaptation is confined to the language model. Gradient
checkpointing is disabled. The same LoRA configuration, data splits, server,
and optimization protocol are used for all four MedAD-R1 backbones. The three
larger SFT comparison models are trained using the same Stage~1 configuration
and the same SFT data.

\subsubsection{Supervised Fine-Tuning}

Stage~1 is trained on the Train-SFT subset of MedAD-38K for two epochs using
six-GPU DDP. The per-device batch size is 8 and gradient accumulation is not
used, yielding a global batch size of 48. The maximum sequence length is
1,024 tokens.

Optimization uses fused AdamW with
$(\beta_1,\beta_2)=(0.9,0.95)$, numerical stability constant
$\epsilon_{\mathrm{Adam}}=10^{-8}$, weight decay 0.1, and maximum gradient
norm 1.0. The initial learning rate is $1\times10^{-4}$ and follows a cosine
decay schedule with a warmup ratio of 0.03.

\subsubsection{Consistency Group Relative Policy Optimization}

Stage~2 starts from the SFT checkpoint and trains on the disjoint Train-RL
subset for one epoch. The per-device prompt batch size is 1 with two gradient
accumulation steps across six GPUs, corresponding to an effective prompt batch
size of 12 per optimization step. Four completions are generated for each
prompt.

All four backbones use the original sample-level GRPO loss
(
\texttt{loss\_type=grpo}
). For each completion, the clipped token-level policy-gradient terms are
averaged over its valid output tokens, after which the completion losses are
averaged across the sampled batch. Rewards are standardized within the four
completions generated for each prompt (\texttt{scale\_rewards=group}), matching
the group-relative advantage in Equation~\ref{eq:advantage}. Each rollout batch
is reused for two policy updates (\texttt{num\_iterations=2}). For every
backbone, the frozen KL-reference policy is the corresponding Stage~1 SFT
checkpoint, while the trainable policy is initialized from the same checkpoint.

Con-GRPO uses fused AdamW with a learning rate of $1\times10^{-5}$, weight
decay 0.1, a cosine schedule, and no warmup. The KL penalty coefficient is
$\beta=0.04$, and the clipping parameter is $\epsilon=0.2$. The maximum input
and completion lengths are 1,536 and 512 tokens, respectively. Rollout
generation uses temperature 1.0, top-$p$ 0.9, and the Transformers backend
without vLLM. Thinking mode is enabled, and the policy is optimized using the
Evidence-Aware Consistency Reward. The fixed MedAD-R1 checkpoint used for the
detailed analyses is trained with seed 42. This is the training seed; no fixed
seed is used for evaluation inference. Accuracy means and standard deviations
are computed from five independent inference passes of this checkpoint and do
not aggregate multiple training runs.

\begin{table}[htbp]
    \centering
    \setlength{\tabcolsep}{8pt}
    \renewcommand{\arraystretch}{1.1}
    \begin{tabular}{lcc}
        \toprule
        \textbf{Setting} & \textbf{Stage 1 (SFT)} &
        \textbf{Stage 2 (Con-GRPO)} \\
        \midrule
        Training subset          & Train-SFT & Train-RL \\
        Epochs                   & 2 & 1 \\
        Training GPUs            & 6 & 6 policy + 2 reward \\
        Per-device batch size    & 8 & 1 \\
        Gradient accumulation    & 1 & 2 \\
        Effective prompt batch   & 48 & 12 \\
        Completions per prompt   & -- & 4 \\
        Learning rate            & $1\times10^{-4}$ & $1\times10^{-5}$ \\
        Optimizer                & Fused AdamW & Fused AdamW \\
        Weight decay             & 0.1 & 0.1 \\
        Scheduler                & Cosine & Cosine \\
        Warmup ratio             & 0.03 & 0 \\
        Maximum input length     & 1,024 & 1,536 \\
        Maximum completion length& -- & 512 \\
        LoRA rank / $\alpha$     & 16 / 32 & 16 / 32 \\
        LoRA dropout             & 0.05 & 0.05 \\
        Precision                & BF16 & BF16 \\
        Loss type                & -- & Original GRPO \\
        Loss normalization       & -- & Per completion length \\
        Reward scaling           & -- & Group \\
        Policy iterations        & -- & 2 \\
        KL reference             & -- & Stage~1 SFT checkpoint \\
        KL coefficient $\beta$   & -- & 0.04 \\
        Clipping $\epsilon$      & -- & 0.2 \\
        Sampling temperature     & -- & 1.0 \\
        Top-$p$                  & -- & 0.9 \\
        Representative seed      & 42 & 42 \\
        \bottomrule
    \end{tabular}
    \caption{Optimization settings for the two MedAD-R1 training stages.}
    \label{tab:hyperparameters}
\end{table}

\subsubsection{Evaluation Deployment}

For the accuracy and reasoning evaluations reported in the main results, all
models are deployed on the same main experimental server using vLLM 0.23.0.
Each fixed checkpoint is evaluated through five independent inference passes
using the default sampling configuration of vLLM 0.23.0: temperature $=1.0$,
top-$p=1.0$, top-$k=0$, $n=1$, and no fixed random seed. The prompts,
answer-option order, test images, and maximum output length remain unchanged
across passes. We report the mean and standard deviation of the five accuracy
results.
This deployment setup is used only for the standard model evaluations. The
Con-GRPO rollouts themselves use the Transformers backend, and the separately
controlled latency and memory experiments use the environment specified in
Section~\ref{app:efficiency}.

\section{Inference Efficiency and Deployment}
\label{app:efficiency}

A small parameter count does not by itself guarantee efficient deployment, since
latency and memory also depend on the number of visual tokens, the architecture,
and the decoding implementation. This section therefore reports measured
inference cost rather than parameter count alone. The results show that MedAD-R1
attains competitive speed at a fraction of the memory required by larger
baselines, which supports its suitability for resource-constrained settings.

\subsection{Measurement Protocol}

All efficiency experiments were conducted on a node equipped with eight NVIDIA
A100-SXM4-80GB GPUs, two Intel Xeon Platinum 8369B processors (64 physical cores
and 128 hardware threads in total), and 800~GiB of system memory. Each model was
evaluated independently on a single A100 GPU without tensor, data, or model
parallelism, while the remaining GPUs were kept idle. The environment ran
Ubuntu~24.04.4 LTS with Python~3.12.3, PyTorch~2.12.1+cu129,
torchvision~0.27.1+cu129, Transformers~5.13.0, CUDA~12.9, and cuDNN~9.2.0. All
models were loaded in bfloat16 precision.

We used a batch size of one and resized all input images to $448\times448$ pixels
with letterbox padding before measurement. For each model, ten samples were used
for warm-up, followed by 500 measured samples. Image loading and resizing were
excluded from the latency measurements, whereas processor computation,
host-to-device transfer, multimodal prefilling, and autoregressive decoding were
included. Generation was deterministic with greedy decoding and KV caching
enabled. We report time to first token (TTFT), time per output token (TPOT),
decoding throughput, end-to-end latency (E2E), output length, and peak allocated
and reserved GPU memory. Because end-to-end latency depends on how many tokens a
model chooses to generate, we report two settings: a fixed-length setting in
which every model produces exactly 256 tokens by setting
\texttt{min\_new\_tokens=256} and \texttt{max\_new\_tokens=256}, which isolates
speed from output length, and a natural setting in which each model stops on
its own, which reflects realistic deployment behavior.

\subsection{Fixed-Length Comparison}

Table~\ref{tab:efficiency_fixed} reports the fixed-length setting, where all
models generate 256 tokens and therefore differ only in speed and memory. Under
this controlled comparison, MedAD-R1 (Qwen3.5-0.8B) is the fastest model, with a
per-token latency of 16.45~ms and a decoding throughput of 60.79~tokens/s, while
completing a full 256-token response in 4.26~s. It is also by far the most
memory-efficient, requiring only 1.82~GB of peak allocated memory. For
comparison, Gemma-4-26B requires 13.24~s and 52.00~GB, and InternVL3.5-8B
requires 6.83~s and 17.18~GB. Across the whole comparison, per-token latency and
peak memory generally increase with model size, and MedAD-R1 sits at the
favorable end of both trends: it uses roughly one order of magnitude less GPU
memory than the larger baselines, supporting its potential for deployment on
resource-constrained hardware.

\begin{table}[htbp]
    \centering
    \small
    \setlength{\tabcolsep}{4pt}
    \begin{tabular}{lccccccc}
        \toprule
        \textbf{Model} & \textbf{Params (B)} & \textbf{TTFT (s)} &
        \textbf{TPOT (ms)} & \textbf{Tokens/s} & \textbf{E2E (s)} &
        \textbf{Peak Alloc. (GB)} & \textbf{Peak Reserv. (GB)} \\
        \midrule
        Lingshu-7B        & 7.0  & 0.0932 & 25.50 & 39.22 & 6.625  & 16.73 & 16.98 \\
        InternVL3.5-8B    & 8.0  & 0.0610 & 26.40 & 37.88 & 6.831  & 17.18 & 17.34 \\
        Qwen3-VL-8B       & 8.0  & 0.2910 & 29.60 & 33.78 & 7.864  & 17.67 & 17.86 \\
        GLM-4.6V-Flash    & 9.0  & 0.0922 & 30.55 & 32.73 & 7.921  & 20.73 & 20.88 \\
        Llama-3.2-Vision  & 11.0 & 0.2845 & 31.20 & 32.05 & 8.267  & 22.30 & 22.90 \\
        Gemma-4-26B       & 26.0 & 0.2261 & 50.83 & 19.67 & 13.238 & 52.00 & 52.21 \\
        \midrule
        MedAD-R1 (InternVL3.5-1B) & 1.0 & 0.0580 & 22.84 & 43.78 & 5.908 & 2.23 & 2.32 \\
        MedAD-R1 (Qwen3VL-2B)     & 2.0 & \textbf{0.0535} & 23.50 & 42.55 & 6.061 & 4.36 & 4.44 \\
        MedAD-R1 (Qwen2.5VL-3B)   & 3.0 & 0.0808 & 24.80 & 40.32 & 6.438 & 8.32 & 8.58 \\
        MedAD-R1 (Ours) & \textbf{0.8} & 0.0585 & \textbf{16.45} & \textbf{60.79} & \textbf{4.260} & \textbf{1.82} & \textbf{1.89} \\
        \bottomrule
    \end{tabular}
    \caption{Inference efficiency under the fixed-length setting (256 output
    tokens). TTFT: time to first token. TPOT: time per output token. E2E:
    end-to-end latency. Peak Alloc.\ and Peak Reserv.\ are peak allocated and
    reserved GPU memory. MedAD-R1 variants are grouped at the bottom. Boldface
    marks the best value in each column.}
    \label{tab:efficiency_fixed}
\end{table}

\subsection{Natural-Length Behavior}

Table~\ref{tab:efficiency_natural} reports the natural setting, where each model
decides its own output length. Here the raw end-to-end latency is not directly
comparable across models, because it is dominated by how many tokens each model
emits. MedAD-R1 (Qwen3.5) produces a concise 49-token answer on average and
finishes in 0.87~s, and all MedAD-R1 variants terminate well before the
generation limit. We include this table to characterize realistic behavior rather
than as a head-to-head speed comparison; the fixed-length results in
Table~\ref{tab:efficiency_fixed} provide the controlled comparison. The memory
footprint is largely insensitive to output length because model weights dominate
the allocation, so the memory advantage of MedAD-R1 holds in both settings.

\begin{table}[htbp]
    \centering
    \small
    \setlength{\tabcolsep}{4pt}
    \begin{tabular}{lccccccc}
        \toprule
        \textbf{Model} & \textbf{Params (B)} & \textbf{TTFT (s)} &
        \textbf{TPOT (ms)} & \textbf{E2E (s)} & \textbf{Out. Tokens} &
        \textbf{Peak Alloc. (GB)} & \textbf{Peak Reserv. (GB)} \\
        \midrule
        Lingshu-7B        & 7.0  & 0.0933 & 25.50 & 3.618  & 138.2  & 16.73 & 16.98 \\
        InternVL3.5-8B    & 8.0  & 0.0612 & 26.40 & 4.925  & 184.3  & 17.22 & 17.34 \\
        Qwen3-VL-8B       & 8.0  & 0.2940 & 29.60 & 5.490  & 175.5  & 17.71 & 17.86 \\
        GLM-4.6V-Flash    & 9.0  & 0.0932 & 30.55 & 15.734 & 512.0$^{*}$ & 20.73 & 20.88 \\
        Llama-3.2-Vision  & 11.0 & 0.2847 & 31.20 & 6.166  & 188.6  & 22.30 & 22.90 \\
        Gemma-4-26B       & 26.0 & 0.2269 & 50.83 & 15.275 & 295.7  & 52.05 & 52.21 \\
        \midrule
        MedAD-R1 (InternVL3.5-1B) & 1.0 & 0.0585 & 22.84 & 1.565 & 66.0  & 2.20 & 2.31 \\
        MedAD-R1 (Qwen3VL-2B)     & 2.0 & \textbf{0.0530} & 23.50 & 2.408 & 100.2 & 4.33 & 4.43 \\
        MedAD-R1 (Qwen2.5VL-3B)   & 3.0 & 0.0808 & 24.80 & 3.602 & 142.0 & 8.32 & 8.58 \\
        MedAD-R1 (Ours) & \textbf{0.8} & 0.0582 & \textbf{16.45} & \textbf{0.866} & \textbf{49.1} & \textbf{1.82} & \textbf{1.89} \\
        \bottomrule
    \end{tabular}
    \caption{Inference behavior under the natural setting, where each model stops
    on its own. End-to-end latency here is governed by output length and is not a
    controlled speed comparison; see Table~\ref{tab:efficiency_fixed} for the
    fixed-length comparison. Boldface marks the lowest numerical value in each
    column; end-to-end latency and output length remain descriptive
    rather than controlled comparisons. $^{*}$GLM-4.6V-Flash reached the maximum generation
    limit of 512 tokens because it did not emit an explicit end-of-sequence token
    in this zero-shot setting.}
    \label{tab:efficiency_natural}
\end{table}

\subsection{Summary}

Across both settings, the primary MedAD-R1 model requires only 1.82~GB of peak
memory, roughly a tenth of what the larger baselines consume, while achieving the
lowest per-token latency and the highest decoding throughput in the comparison.
Combined with the accuracy and consistency results in Section~\ref{app:extended},
this indicates that the advantages of MedAD-R1 are obtained without any additional
inference burden and supports its potential for deployment on
resource-constrained hardware.

\FloatBarrier
\section{Qualitative Comparisons}
\label{app:qualitative}

This section presents reasoning traces generated during evaluation. The quoted
passages are verbatim excerpts from the recorded model outputs, and the
corresponding ground-truth answers are taken from the source annotations.

\subsection{Cases Where MedAD-R1 Improves Over the Baselines}
\label{app:qualitative_success}

We select two cases for which the same six comparison models are available:
one liver-mass localization case and one thyroid abnormality-detection case.
For MedAD-R1, we show only the primary Qwen3.5-0.8B backbone. The comparison
set contains two large general-purpose LMMs and four domain-specific medical
LMMs. These are selected success cases rather than an estimate of aggregate
performance. All answers and reasoning excerpts are reproduced from the
recorded evaluation outputs; wrapper tags are omitted and ``\ldots'' indicates
omitted text. We manually inspected each image before inclusion, and both cases
provide clear visual evidence for the source-derived answers.

\begingroup
\setlength{\textfloatsep}{8pt plus 2pt minus 2pt}
\setlength{\intextsep}{8pt plus 2pt minus 2pt}
\setlength{\tabcolsep}{4pt}
\renewcommand{\arraystretch}{1.08}
\newcommand{\modeloutput}[2]{\par\smallskip\noindent\textbf{#1}\quad #2\par}

\begin{table}[htbp]
    \centering
    \begin{tabular}{lcccccccc}
        \toprule
        \textbf{Case} & \textbf{Gold} & \textbf{Ours} &
        \textbf{InternVL} & \textbf{Gemma} & \textbf{Huatuo$^{*}$} &
        \textbf{LLaVA$^{*}$} & \textbf{Lingshu$^{*}$} &
        \textbf{MedVLM$^{*}$} \\
        \midrule
        Liver-mass location & B & B & A & A & A & C & C & C \\
        Thyroid finding     & B & B & A & A & A & A & A & A \\
        \bottomrule
    \end{tabular}
    \caption{Answer-level summary of the two manually reviewed qualitative
    cases. $^{*}$ denotes a domain-specific medical LMM. All 12 comparison
    answers are incorrect under the source-derived labels.}
    \label{tab:qualitative_case_answers}
\end{table}

\paragraph{Case 1: Organ location substituted for lesion location.}
Figure~\ref{fig:qual_liver_mass} contains a large heterogeneous liver mass. The
source label is determined by the five-region rule applied to the segmentation
mask, not by the conventional anatomical quadrant occupied by the liver.

\begin{figure}[htbp]
    \centering
    \includegraphics[width=0.25\textwidth]
    {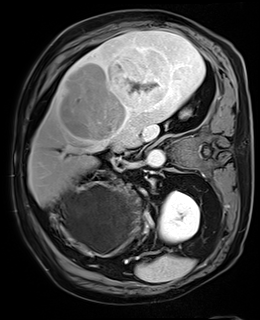}
    \caption{Lesion-localization example (source-derived answer: B,
    Center-Region). The annotated finding spans a large portion of the liver and
    is assigned to the center region by the mask-based five-region rule.}
    \label{fig:qual_liver_mass}
\end{figure}

\noindent\textbf{Question.}
An anomaly has been identified in this image. Where is it precisely located?
\emph{A: Right-Upper; B: Center-Region; C: Right-Lower; D: Left-Upper.}

\modeloutput{MedAD-R1 (Ours); answer B.}{The image shows a large,
heterogeneous mass located centrally within the liver. It occupies a
significant portion of the liver's core region. \ldots{} The lesion's location
is best described as the center region of the liver.}

\modeloutput{InternVL3.5-38B; answer A.}{The liver is situated in the upper
right part of the abdomen. The mass is clearly visible in this region.
\ldots{} The mass is located in the upper right part of the image, which
corresponds to the right-upper region of the abdomen.}

\modeloutput{Gemma-4-26B; answer A.}{There is a large, prominent mass/lesion
occupying a significant portion of the upper abdomen. \ldots{} While it extends
toward the center, its primary location is the upper right.}

\modeloutput{HuatuoGPT-Vision$^{*}$; answer A.}{The image provided is a
cross-sectional view of the abdomen, likely from an MRI scan. The anomaly
appears to be located in the upper right quadrant of the abdomen.}

\modeloutput{LLaVA-Med$^{*}$; answer C.}{The image shows a cross-sectional view
of the abdomen. The anomaly is located in the right lower quadrant of the
image. Therefore, the correct answer is C) Right-Lower.}

\modeloutput{Lingshu-7B$^{*}$; answer C.}{Upon close inspection, there is a
noticeable mass-like structure in the lower right quadrant of the image. This
area corresponds to the anatomical region known as the ``Right-Lower.''}

\modeloutput{MedVLM-R1$^{*}$; answer C.}{The anomaly appears to be located in
the right lower quadrant of the abdomen. This is because the right lower
quadrant is typically the location of the kidneys, which are visible in the
image. Therefore, the anomaly is most likely located in the right lower
quadrant.}

\noindent\emph{Analysis.}
The three Right-Upper answers transfer the liver's conventional abdominal
location, or the right-lobe description, to a label defined by the lesion mask
in image coordinates. The other three models assert a lower-right focus;
MedVLM-R1 further relies on the visibility of a kidney, which is unrelated to
the mass location. MedAD-R1 instead describes the lesion's central extent
within the liver and selects the mask-derived Center-Region label.

\FloatBarrier

\paragraph{Case 2: A visible thyroid finding described as absent.}
Figure~\ref{fig:qual_thyroid_one} shows a focal heterogeneous hypoechoic area in
the central thyroid region.

\begin{figure}[htbp]
    \centering
    \includegraphics[width=0.28\textwidth]
    {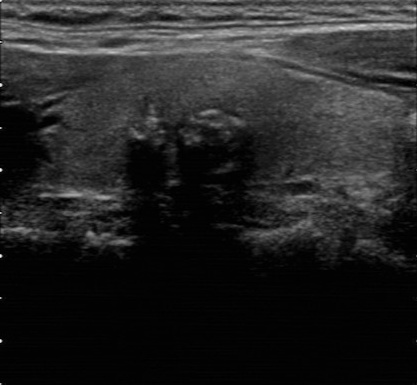}
    \caption{Thyroid-ultrasound anomaly-detection example (source-derived
    answer: B, Yes). A focal echotextural abnormality is distinguishable from
    the surrounding thyroid parenchyma.}
    \label{fig:qual_thyroid_one}
\end{figure}

\noindent\textbf{Question.}
Are there any clinically significant findings visible?
\emph{A: No; B: Yes.}

\modeloutput{MedAD-R1 (Ours); answer B.}{The image shows a focal area within
the thyroid gland that differs in echotexture from the surrounding tissue.
This area appears hypoechoic and has somewhat irregular borders, suggesting a
nodule or lesion.}

\modeloutput{InternVL3.5-38B; answer A.}{There are no obvious signs of nodules,
cysts, or other abnormalities in the thyroid tissue. The echotexture appears
homogeneous, and there are no areas of significant hypoechogenicity or
hyperechogenicity that would suggest pathology.}

\modeloutput{Gemma-4-26B; answer A.}{The provided image is a low-resolution
ultrasound scan. \ldots{} There are no clear, identifiable pathological
features visible to a trained eye. Therefore, based strictly on what is visible
in this specific image, no significant findings can be identified.}

\modeloutput{HuatuoGPT-Vision$^{*}$; answer A.}{As an AI, I am not capable of
providing a medical diagnosis or identifying clinically significant findings
in medical images. For an accurate interpretation, please consult a qualified
healthcare professional.}

\modeloutput{LLaVA-Med$^{*}$; answer A.}{The image shows a structure that
appears to be a kidney. However, without clinical context or additional
information, it is not possible to determine if there are any clinically
significant findings.}

\modeloutput{Lingshu-7B$^{*}$; answer A.}{The image appears to show a
homogeneous echotexture without any obvious masses, cysts, or irregularities
that might indicate pathology. \ldots{} The absence of any hypoechoic or
hyperechoic regions suggests that there is no evidence of lesions or
significant abnormalities.}

\modeloutput{MedVLM-R1$^{*}$; answer A.}{The ultrasound image shows a
normal-looking organ, but without additional clinical context, it is difficult
to make a definitive diagnosis. Therefore, the most appropriate answer is A,
No, as there are no clinically significant findings visible.}

\noindent\emph{Analysis.}
The focal heterogeneous hypoechoic area and associated acoustic change are
visible in the central thyroid region. InternVL3.5-38B, Lingshu-7B, and
MedVLM-R1 describe the image as homogeneous or normal; Gemma-4-26B dismisses
the image as insufficient; HuatuoGPT-Vision declines the requested task; and
LLaVA-Med misidentifies the anatomy. MedAD-R1 identifies the visible focal
finding while avoiding a specific histopathologic diagnosis.

\FloatBarrier
\section{Open-source release}
\label{app:Open_source}
Upon publication, we will release the project-owned annotations, metadata,
source code, evaluation scripts, and model weights to the extent permitted by
the licenses of the underlying datasets and base models. Source code will be
released under the MIT License, while model weights will follow the applicable
base model license.

\endgroup